\documentclass[10pt,twocolumn,letterpaper]{article}

\usepackage{cvpr}
\usepackage{times}
\usepackage{epsfig}
\usepackage{graphicx}
\usepackage{amsmath}
\usepackage{amssymb}
\usepackage{balance}

\usepackage[format=plain,labelformat=simple,labelsep=period,font=small,skip=4pt,compatibility=false]{caption}
\usepackage[font=footnotesize,skip=2pt]{subcaption}
\newcommand{\myparagraph}[1]{\medbreak\noindent\textbf{#1}}

\usepackage{pifont}
\newcommand{\cmark}{\ding{51}}

\usepackage{booktabs}
\usepackage{multirow}
\usepackage{tabularx}
\usepackage{visinfdefs}
\usepackage[super]{nth}

\usepackage{etoolbox}
\usepackage[binary-units]{siunitx}
\robustify\bfseries
\DeclareSIUnit{\inch}{inch}

\usepackage{tikz}
\newcommand*\circled[1]{\tikz[baseline=(char.base)]{
        \node[shape=circle,draw,fill=white,minimum size=3mm,inner sep=0pt] (char)
        {\vphantom{1g}\emph{#1}};}}
        
\usepackage[breaklinks=true,bookmarks=false,colorlinks,pagebackref=true]{hyperref}
\usepackage[capitalize]{cleveref}
\crefname{section}{Sec.}{Section}

\newcommand{\supptitle}{Iterative Residual Refinement for Joint Optical Flow and Occlusion Estimation \\ {\large -- Supplementary Material --}}
	\newcommand{\suppauthor}{Junhwa Hur \qquad\qquad Stefan Roth \\
Department of Computer Science, TU Darmstadt}

\cvprfinalcopy 
\def\cvprPaperID{****} 

\ifcvprfinal\fi

\usepackage{fancyhdr}
\usepackage{setspace}

\begin{document}
\title{Iterative Residual Refinement for Joint Optical Flow and Occlusion Estimation}

\author{Junhwa Hur \qquad\qquad Stefan Roth \\
Department of Computer Science, TU Darmstadt}

\maketitle
\thispagestyle{fancy} 

\begin{abstract}
Deep learning approaches to optical flow estimation have seen rapid progress over the recent years.
One common trait of many networks is that they refine an initial flow estimate either through multiple stages or across the levels of a coarse-to-fine representation.
While leading to more accurate results, the downside of this is an increased number of parameters.
Taking inspiration from both classical energy minimization approaches as well as residual networks, we propose an \emph{iterative residual refinement} (IRR) scheme based on \emph{weight sharing} that can be combined with several backbone networks.
It reduces the number of parameters, improves the accuracy, or even achieves both.
Moreover, we show that integrating occlusion prediction and bi-directional flow estimation into our IRR scheme can further boost the accuracy.
Our full network achieves state-of-the-art results for both optical flow and occlusion estimation across several standard datasets.
\end{abstract}


\section{Introduction}
\label{sec:introduction}


Akin to many areas of computer vision, deep learning has had a significant impact on optical flow estimation.
But in contrast to, \eg, object detection \cite{Girshick:2016:RBC} or human pose estimation \cite{Tompson:2014:JTC}, the accuracy of deep learning-based flow methods on public benchmarks \cite{Butler:2012:NOS,Geiger:2012:AWR,Menze:2015:OSF} had initially not surpassed that of classical approaches.
Still, the efficient test-time inference has led to their widespread adoption as a sub-module in applications requiring to process temporal information, including video object segmentation \cite{Cheng:2017:SFJ}, video recognition \cite{Gadde:2017:SVC,Nilsson:2018:SVS,Zhu:2017:DFF}, and video style transfer \cite{Chen:2017:COV}.

FlowNet \cite{Dosovitskiy:2015:FLO} pioneered the use of convolutional neural networks (CNNs) for estimating optical flow and relied on a -- by now standard -- encoder-decoder architecture with skip connections, similar to semantic segmentation \cite{Long:2015:FCN}, among others.
Since the flow accuracy remained behind that of classical methods based on energy minimization, later work has focused on designing more powerful CNN architectures for optical flow.
FlowNet2 \cite{Ilg:2017:FN2} remedied the accuracy limitations of FlowNet and started to outperform classical approaches.
Its main principle is to stack multiple FlowNet-family networks \cite{Dosovitskiy:2015:FLO}, such that later stages effectively refine the output from the previous ones.
However, one of the side effects of this stacking is the linearly and strongly increasing number of parameters, being a burden for the adoption in other applications.
Also, stacked networks require training the stages sequentially rather than jointly, resulting in a complex training procedure in practice.

\begin{figure}[t]
\centering 
\includegraphics[width=0.9\linewidth]{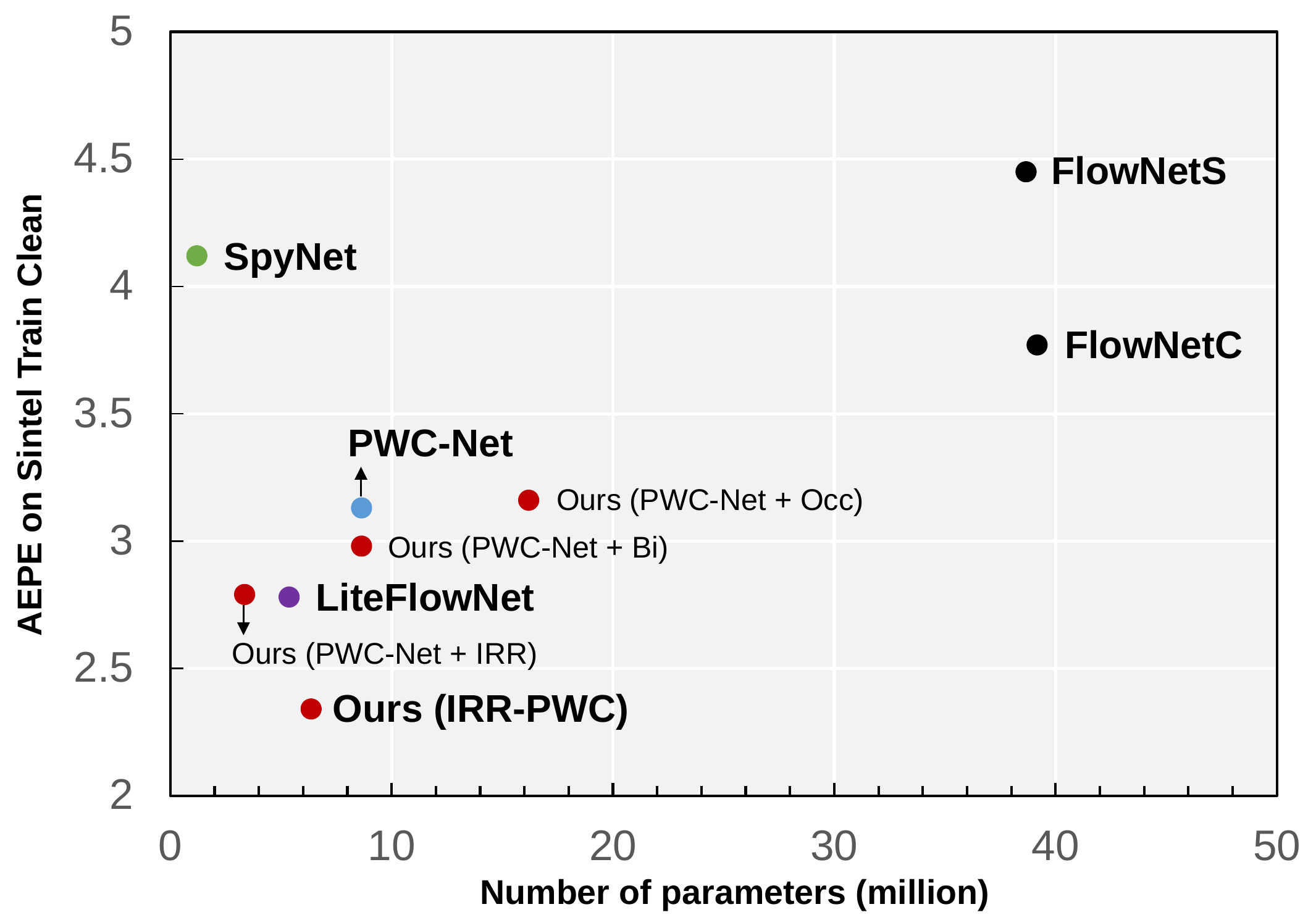}
\caption{\textbf{Accuracy / network size tradeoff of CNNs for optical flow:}
Combining our iterative residual refinement (IRR), as well as bi-directional (Bi) and occlusion estimation (Occ) with PWC-Net \cite{Sun:2017:PWC} in comparison to previous work.
Our full model (IRR-PWC), combining all three components, yields significant accuracy gains over \cite{Sun:2017:PWC} while having many fewer parameters.}
\label{fig:param_plot}
\vspace{-0.5em}
\end{figure}

More recently, SpyNet \cite{Ranjan:2017:OFE}, PWC-Net \cite{Sun:2017:PWC}, and LiteFlowNet \cite{Hui:2018:LFN} proposed lightweight networks that still achieve competitive accuracy (\cf\cref{fig:param_plot}).
SpyNet adopts coarse-to-fine estimation in the network design, a well-known principle in classical approaches.
It residually updates the flow across the levels of a spatial pyramid with individual trainable weights and demonstrates better accuracy than FlowNet but with far fewer model parameters.
LiteFlowNet and PWC-Net further combine the coarse-to-fine strategy with multiple ideas from both classical methods and recent deep learning approaches. 
Particularly PWC-Net outperformed all published methods on the common public benchmarks \cite{Butler:2012:NOS,Geiger:2012:AWR,Menze:2015:OSF}.

\begin{figure*}[t]
\centering
\subcaptionbox{FlowNet2-like \cite{Ilg:2017:FN2} stack of FlowNet networks.\label{fig:intro_a}}{\includegraphics[width=0.435\textwidth]{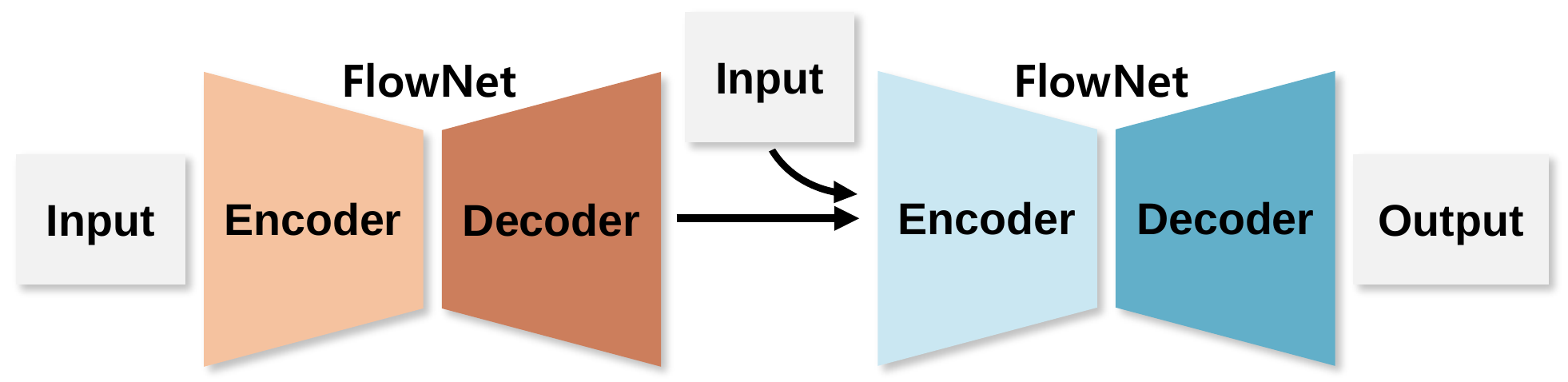}} \quad
\subcaptionbox{Iterative residual refinement version of \emph{(\subref{fig:intro_a})}.\label{fig:intro_b}}{\includegraphics[width=0.49\textwidth]{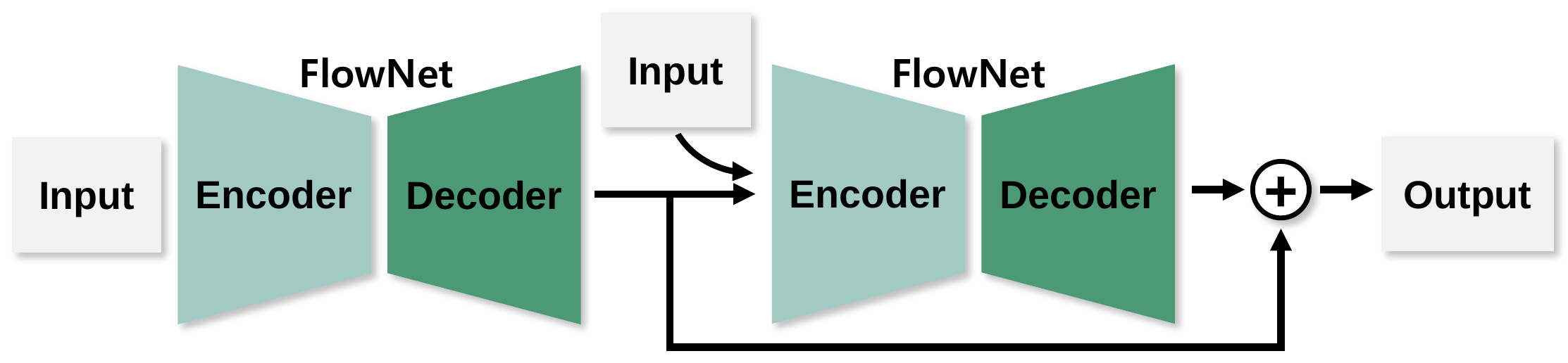}} \\[0.5mm]
\subcaptionbox{A rolled version of \emph{(\subref{fig:intro_b})}.\label{fig:intro_c}}{\includegraphics[width=0.34\textwidth]{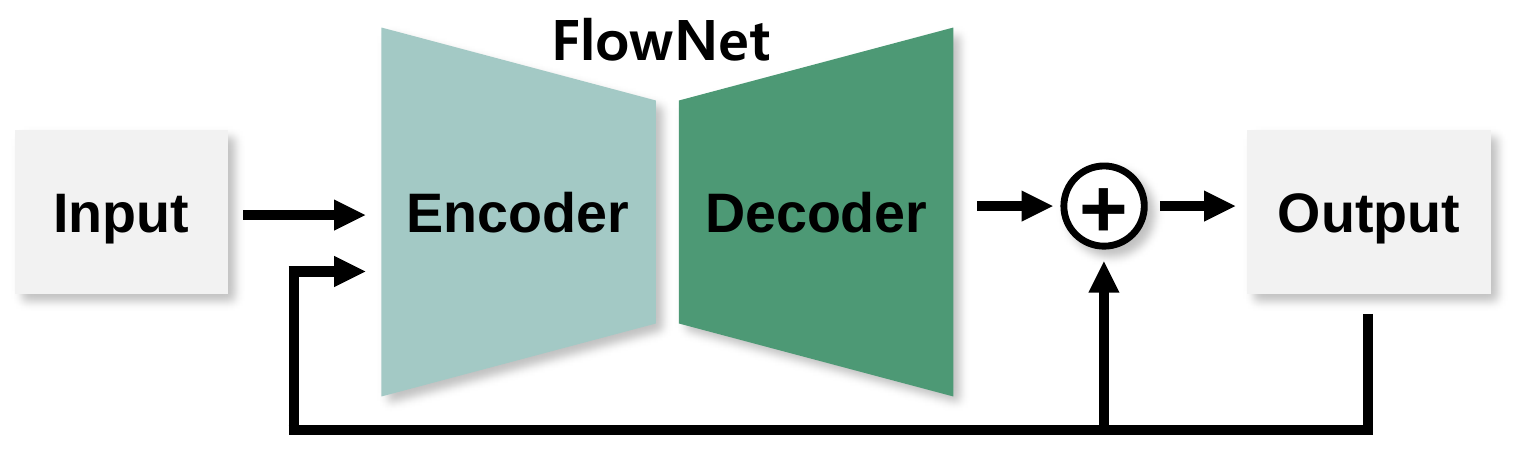}} \hspace{2cm}
\subcaptionbox{Joint flow and occlusion estimation.\label{fig:intro_d}}{\includegraphics[width=0.335\textwidth]{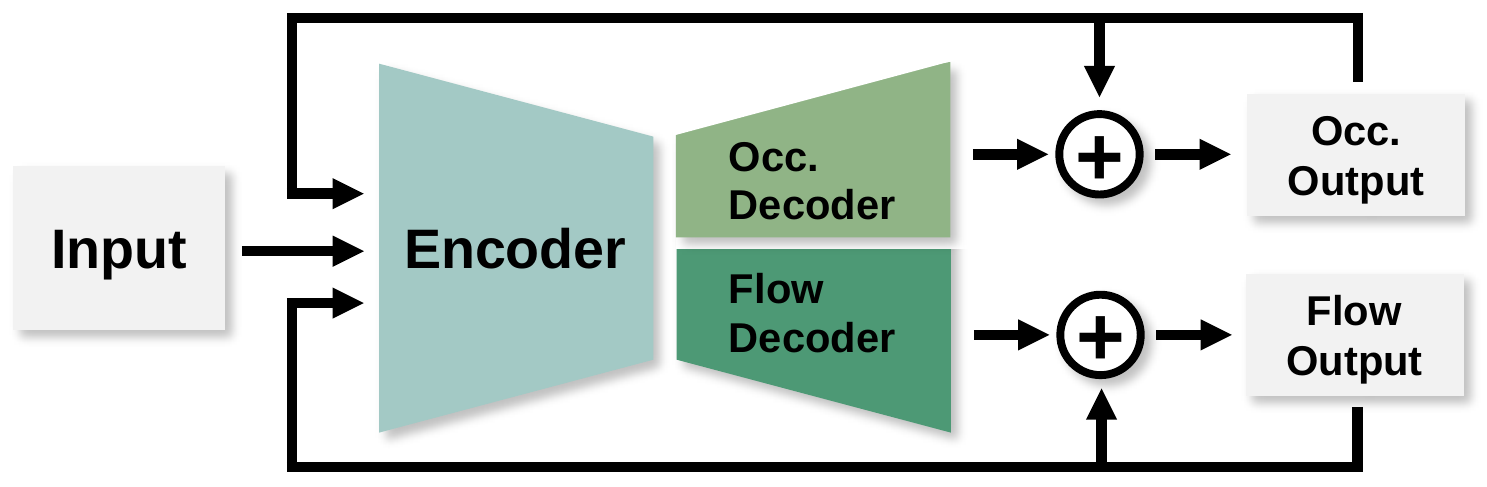}} \hspace{0.9cm}
\caption{\textbf{From a standard network stack to our iterative residual refinement scheme with joint optical flow and occlusion estimation:} 
The stacked version of FlowNet \emph{(\subref{fig:intro_a})} can be converted into an iterative residual refinement model \emph{(\subref{fig:intro_b})} with the half number of parameters.
Note that modules with the same color share their weights.
We can re-interpret \emph{(\subref{fig:intro_b})} as a rolled version \emph{(\subref{fig:intro_c})}, making it more immediate to include occlusion estimation \emph{(\subref{fig:intro_d})} for further improving the accuracy.}
\label{fig:intro}
\vspace{-0.5em}
\end{figure*}

Interestingly, many recent deep learning approaches for flow \cite{Hui:2018:LFN,Ilg:2017:FN2,Ranjan:2017:OFE,Sun:2017:PWC} have a common structure:
From a rough first flow estimate, later modules or networks refine the previous estimates across pyramid levels or through multiple chained networks.
As illustrated in \cref{fig:intro_a}, the later modules or networks have their own trainable weights, since each module assumes a particular functionality at the respective spatial resolution or conditioned on the output of the preceding modules.
The downside is that this significantly increases the number of required model parameters.

In this paper, we take the inspiration from classical energy minimization-based optical flow approaches several steps further.
Energy-based methods iteratively estimate the flow based on a consistent underlying energy with a single set of parameters \cite{Black:1996:TRE,Brox:2004:HAO,Sun:2014:QAC}.
Hence we ask: \emph{Can we iteratively refine flow with a deep network based on a single, shared set of weights?}
Moreover, energy-based methods have benefited from bi-directional estimation and occlusion reasoning \cite{Alvarez:2007:SDO,Hur:2017:MFE,Sun:2014:LLJ,Xiao:2006:BFO}.
We thus ask: \emph{Can deep learning approaches to optical flow similarly benefit from bi-directional estimation with occlusion reasoning?}

We address these questions and make a number of contributions:
\emph{(i)} We first propose an iterative residual refinement (IRR) scheme that takes the output from a previous iteration as input and iteratively refines it by only using a \emph{single} network block with \emph{shared weights}.
\emph{(ii)} We demonstrate the applicability to two popular networks, FlowNet \cite{Dosovitskiy:2015:FLO} (\cref{fig:intro_c}) and PWC-Net \cite{Sun:2017:PWC} (\cref{fig:pwcnet}).
For FlowNet, we can significantly increase the accuracy without adding parameters; for PWC-Net, we can reduce the number of parameters while even improving the accuracy (\cref{fig:param_plot}).
\emph{(iii)} Next, we demonstrate the integration with occlusion estimation (\cref{fig:intro_d}).
\emph{(iv)} We further extend the scheme to bi-directional flow estimation, which turns out to be only beneficial when combined with occlusion estimation.
Unlike previous work \cite{Ilg:2018:OMD}, our scheme enables the flow accuracy to benefit from joint occlusion estimation.
\emph{(v)} We finally propose lightweight bilateral filtering and occlusion upsampling layers for refined motion and occlusion boundaries.

Applying our proposed scheme to two backbone networks, FlowNet and PWC-Net, yields significant improvements in flow accuracy of $18.5\%$ and $17.7\%$, respectively, across multiple datasets.
In case of PWC-Net, we achieve this accuracy gain using $26.4\%$ \emph{fewer} parameters.
Note that occlusion estimation and bi-directional flow are additional outcomes as by-products of this improvement.




\section{Related Work}
\label{sec:relatedwork}

\paragraph{Optical flow with CNNs.}
Starting with FlowNet \cite{Dosovitskiy:2015:FLO}, various deep network architectures for optical flow have been proposed, \eg,~FlowNet2 \cite{Ilg:2017:FN2}, SpyNet \cite{Ranjan:2017:OFE}, PWC-Net \cite{Sun:2017:PWC}, and LiteFlowNet \cite{Hui:2018:LFN}. 
They are based on an autoencoder design, allow for supervised end-to-end training, and enable fast inference during testing time. 
To alleviate the need for training data with ground truth in a specific domain, unsupervised \cite{Ahmadi:2016:UCN,Meister:2018:ULO,Ren:2017:UDL,Wang:2018:OAU,Yu:2016:BBU,Zhu:2017:DDF} and semi-supervised \cite{Lai:2017:SSL,Zhu:2017:GOF} alternatives have also been developed.

Other than such end-to-end approaches, CNNs can also serve to extract learned feature descriptors, which are combined with classical optimizers or well-designed correspondence search methods to find matches between extracted features \cite{Bai:2016:ESI,Bailer:2017:CPM,Gadot:2016:PBB,Gueney:2016:DDF,Wulff:2017:OFM,Xu:2017:AOF}. 
Such optimization-based approaches can yield less blurry results than typical CNN decoders, but not all are end-to-end trainable and their runtime in the testing phase is significantly longer. 

Here, we investigate how to improve generic auto\-encoder-based architectures by adapting an iterative residual scheme that is widely applicable.

\begin{figure*}[t]
\centering     
\includegraphics[width=0.72\textwidth]{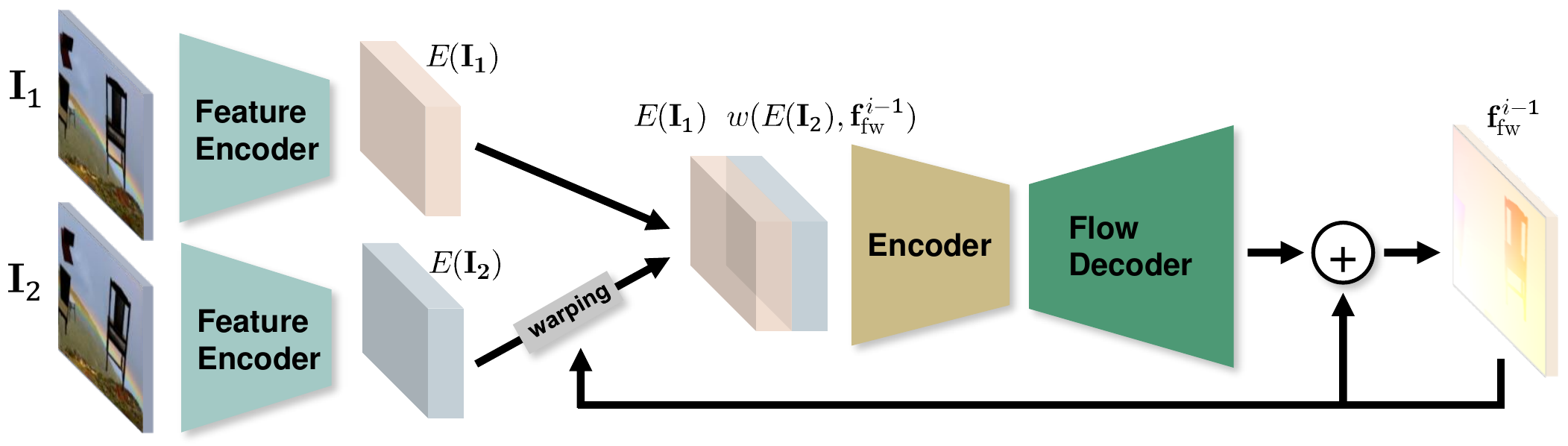}
\caption{\textbf{Our Iterative Residual Refinement (IRR) version of FlowNetS} \cite{Dosovitskiy:2015:FLO}\textbf{.} 
The model iteratively estimates residual flow from the previous output.
Note that we apply warping after several encoder layers, see text for details.}
\label{fig:flownet_iterres}
\vspace{-0.5em}
\end{figure*}

\myparagraph{Optical flow and occlusion.}
Occlusion has been regarded as an important cue for estimating more accurate optical flow.
Because occluded pixels do not have correspondences in the other frame, several approaches \cite{Bailer:2015:FFD,Chen:2016:FFO,Gadot:2016:PBB,Hu:2016:ECF,Li:2015:SSP} aim to filter out these outliers to minimize their ill effects and apply post-processing to refine the estimates \cite{Hu:2017:RIC,LI:2016:FGG,Revaud:2015:EEP,Zweig:2017:INA}.

Other methods \cite{Alvarez:2007:SDO,Ballester:2012:ATL,Hur:2017:MFE,Ince:2008:OOF,Sun:2014:LLJ,Unger:2012:JME,Xiao:2006:BFO} model occlusions in a joint energy and utilize them as additional evidence for flow through recursive joint estimation.
Occlusion estimates can enable \emph{(i)} a more accurate matching cost \cite{Sun:2014:LLJ,Unger:2012:JME,Xiao:2006:BFO}, \emph{(ii)} bi-directional consistency \cite{Hur:2017:MFE,Ince:2008:OOF}, or \emph{(iii)} uniqueness constraints for pixel-level matching \cite{Alvarez:2007:SDO,Hur:2017:MFE,Sun:2005:SSM}.

Recently, \cite{Wang:2018:OAU} proposed an unsupervised deep network that jointly estimates flow and occlusion.
Occlusions are explicitly detected from the inverse of disocclusion \cite{Hur:2017:MFE}; the per-pixel loss is disabled on occluded pixels. 
\cite{Ilg:2018:OMD} proposed a supervised network for jointly estimating optical flow and occlusion, as well as depth and motion boundaries.
\cite{Janai:2018:ULM,Neoral:2018:COO} integrate occlusion estimation into a PWC-Net backbone based on temporal propagation in longer sequences; \cite{Janai:2018:ULM} is based on unsupervised learning.
Our work also directly learns to estimate occlusion using ground-truth supervision signals, but requires only two frames and unlike \cite{Ilg:2018:OMD} enables to improve the flow using the estimated occlusions.

\myparagraph{Iterative and residual refinement.}
Despite of the immense learning capacity of deep networks, early CNN approaches to optical flow did not outperform classical methods \cite{Dosovitskiy:2015:FLO}.
Often, the CNN decoder yielded blurry, thus less accurate regression results.
In order to overcome this and motivated by classical coarse-to-fine refinement \cite{Black:1996:TRE,Brox:2004:HAO,Sun:2014:QAC}, SpyNet \cite{Ranjan:2017:OFE} and PWC-Net \cite{Sun:2017:PWC} residually update the flow along a pyramid structure. 
FlowNet2 \cite{Ilg:2017:FN2} instead stacks multiple networks to refine the previous estimates, though this linearly increases the network size.
DeMoN \cite{Ummenhofer:2017:DDM} uses a combined strategy of stacking and iteratively using one network, also requiring more parameters than one single baseline network.
LiteFlowNet \cite{Hui:2018:LFN} cascades extra convolution layers for refining the outputs and regularizes the outputs based on a feature-driven local convolution, which adaptively defines the convolution weights based on the estimated outputs. 
Related approaches have also been proposed in the stereo literature \cite{Gidaris:2017:DRR,Liang:2018:LDE,Pang:2017:CRL}, successfully improving the accuracy but still increasing the network size.

In contrast, we propose a generic scheme that repetitively uses one baseline network to yield better accuracy without increasing the network size.
For certain networks, we even reduce the size by removing repetitive modules while still enabling competitive or even improved accuracy.

\section{Approach}
\label{sec:approach}


\subsection{Core concepts \& base networks}
\paragraph{Iterative residual refinement (IRR) with shared weights.}
The basic problem setup is to estimate (forward) optical flow $\mv{f}_\text{fw}$ from the reference frame $\mv{I}_1$ to the target frame $\mv{I}_2$.
The main concept of our IRR scheme is to make a model learn to residually refine its previous estimate by iteratively \emph{re-using} the same network block with \emph{shared weights}. 
We pursue two scenarios: \textit{(i)} We increase the accuracy without adding parameters or complicating the training procedure, by iteratively re-using a single network to keep refining its previous estimate; or \textit{(ii)} we aim toward a more compact model by substituting multiple network blocks assuming the same basic functionality with only a single block.

\myparagraph{IRR with FlowNet.}
Addressing the first scenario, we propose an iterative residual refinement version of FlowNetS \cite{Dosovitskiy:2015:FLO}, \cf\cref{fig:flownet_iterres} for an overview.
Our IRR version iteratively estimates residual flow with multiple iterations using one single FlowNetS; the final result is the sum of residual flows from all iteration steps.
We use one shared encoder $E$ for feature extraction from each input image $\mv{I}_1$ and $\mv{I}_2$, similar to FlowNetC, and concatenate the two feature maps after warping the second feature map based on the estimated flow $\mv{f}_\text{fw}^{i-1}$ from the previous iteration $i-1$.
Then we input the concatenated feature maps to the decoder $D$ to estimate the residual (forward) flow at iteration $i$:
\begin{equation}
\mv{f}_\text{fw}^i = D\Big(E(\mv{I}_1), w\big(E(\mv{I}_2), \mv{f}_\text{fw}^{i-1}\big)\Big) + \mv{f}_\text{fw}^{i-1}, 
\label{eq:flownet_baseline}
\end{equation}
where $w(\cdot, \cdot)$ is a bilinear interpolation function for backward warping \cite{Jaderberg:2015:STN}.
Here, warping the second feature map is crucial as it yields a suitable input for estimating the appropriate \emph{residual} flow.
This yields much improved accuracy while re-using the same network with only slight modifications and not requiring additional training stages.

\begin{figure}[t]
\centering     
\subcaptionbox{Original PWC-Net \cite{Sun:2017:PWC}.\label{fig:pwcnet_standard}}{\includegraphics[width=\linewidth]{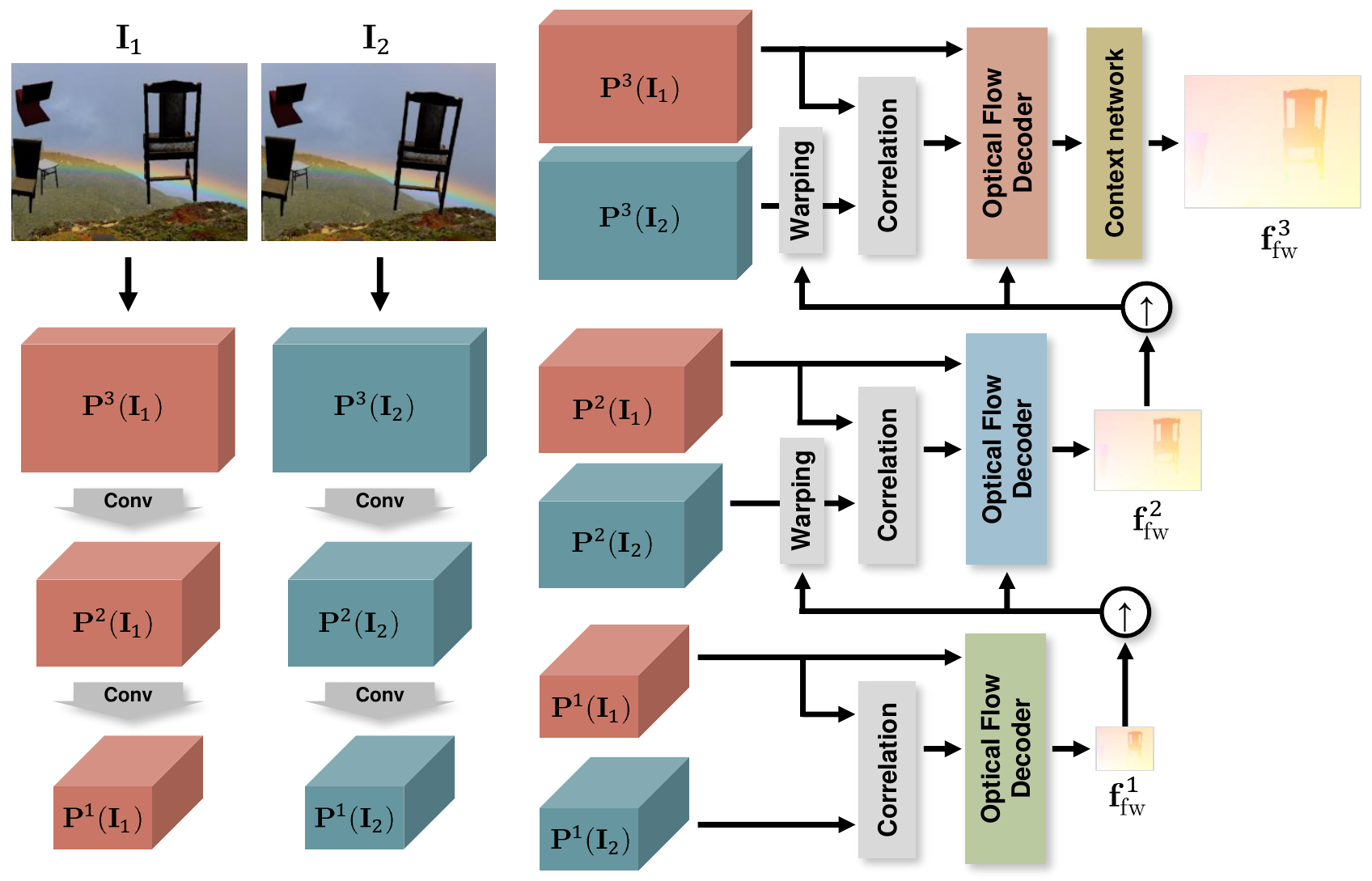}} \\[1mm]
\subcaptionbox{Our IRR version of PWC-Net.\label{fig:pwcnet_shared}}{\includegraphics[width=0.93\linewidth]{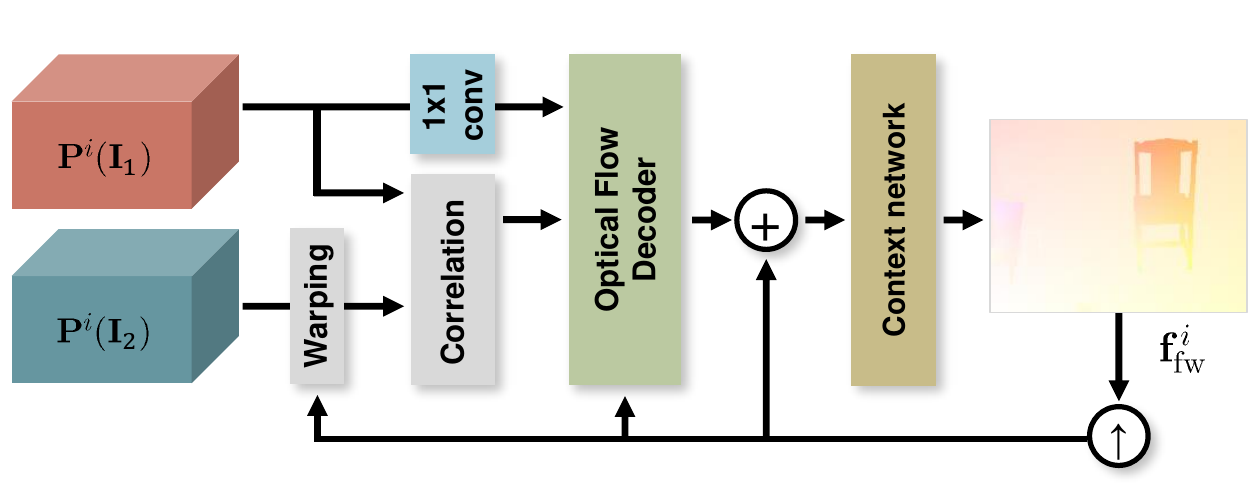}}
\caption{\textbf{Our IRR version of PWC-Net}, which uses only one single shared decoder over the pyramid levels, see text for details.}
\label{fig:pwcnet}
\vspace{-0.8em}
\end{figure}

\myparagraph{IRR with PWC-Net.}
Based on the classical coarse-to-fine principle, PWC-Net \cite{Sun:2017:PWC} and SpyNet \cite{Ranjan:2017:OFE} both use multiple repetitive modules for the same purpose but with \emph{separate weights}.
\cref{fig:pwcnet_standard} shows a 3-level PWC-Net (for ease of visualization, originally 7-level) that incrementally updates the estimation across the pyramid levels with individual decoders for each level.
Adopting our IRR scheme here to address the second scenario, we can substitute the multiple decoders with only one shared decoder that iteratively refines the output over all the pyramid levels, \cf\cref{fig:pwcnet_shared}.
We set the number of iterations equal to the number of pyramid levels, keeping the original pipeline but with fewer parameters and a more compact representation:
\begin{subequations}
\begin{equation}
\label{eq:pwc_net}
\mv{f}_\text{fw}^i = D\Big(\mv{P}^i(\mv{I}_1), c\big(\mv{P}^i(\mv{I}_1), w\big(\mv{P}^i(\mv{I}_2), \hat{\mv{f}}_\text{fw}^{i-1}\big)\big), \hat{\mv{f}}_\text{fw}^{i-1}\Big) + \hat{\mv{f}}_\text{fw}^{i-1}  
\end{equation}
with
\begin{equation}
\label{eq:pwc_net_flow_up}
\hat{\mv{f}}_\text{fw}^{i-1} = 2 \cdot \uparrow\!(\mv{f}_\text{fw}^{i-1}),
\end{equation}
\end{subequations}
where $\mv{P}^i$ is the feature map at pyramid level $i$, $c(\cdot, \cdot)$ calculates a cost volume, and $\uparrow$ performs $2\times$ bilinear upsampling to twice the resolution of the previous flow field.
As the dimension increases, we also scale the flow magnitude accordingly (\Eq\ref{eq:pwc_net_flow_up}).

One important change from the original PWC-Net \cite{Sun:2017:PWC}, which estimates flow for each level on the original scale, is that we estimate flow for each level at its native spatial resolution.
This enables us to use only one shared decoder and yet make it possible to handle different resolutions across all levels.
When calculating the loss, we revert back to the original scale to use the same loss function.

In addition, we add a $1\times1$ convolution layer after the input feature map $\mv{P}^i(\mv{I}_1)$ to make the number of feature maps input to the decoder $D$ be equal across the pyramid levels. 
This enables us to use one single shared decoder with a fixed number of input channels across the pyramid.

\myparagraph{Occlusion estimation.}
It is widely reported that jointly localizing occlusions and estimating optical flow can benefit each other \cite{Alvarez:2007:SDO,Ballester:2012:ATL,Hur:2017:MFE,Ince:2008:OOF,Sun:2014:LLJ,Unger:2012:JME,Xiao:2006:BFO}.
Toward leveraging this in the setting of CNNs, we attach an additional decoder estimating occlusion $\mv{o}_1^i$ in the first frame at the end of the encoder, in parallel to the flow decoder as shown in \cref{fig:intro_d}, similar to \cite{Janai:2018:ULM, Neoral:2018:COO}. 
The occlusion decoder has the same configuration as the flow decoder, but the number of output channels is $1$ (instead of $2$ for flow). 
The input to the occlusion decoder is the same as to the flow decoder.

\subsection{Joint optical flow and occlusion estimation}

\begin{figure*}[t]
\centering
\begin{minipage}{0.73\textwidth}
	\centering
	\includegraphics[width=\linewidth]{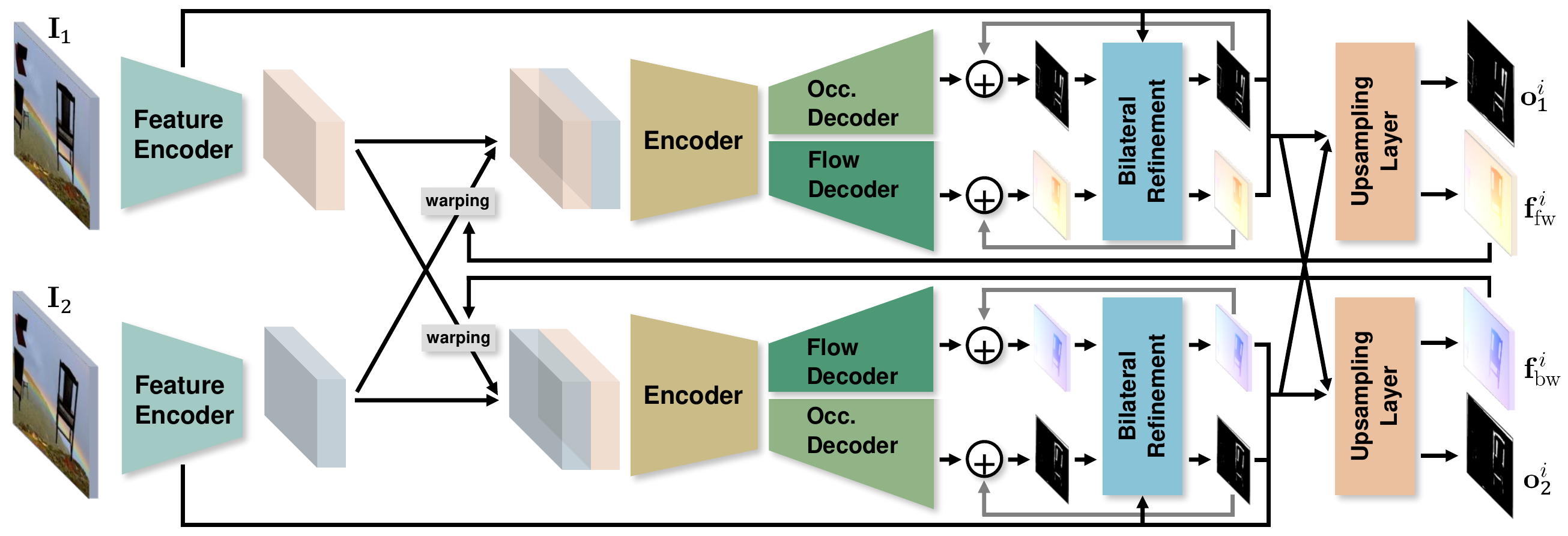}
	\caption{\textbf{Joint optical flow and occlusion estimation: bi-directional estimation, bilateral refinement, and upsampling layer} (in the FlowNet setting): We estimate flow in both temporal directions and occlusion maps in both frames by switching the order of inputs to the decoder. Bilateral refinement and the upsampling layer further improve the accuracy of flow and occlusion. Modules with the same color share their weights.
	\Cf supplemental material for the corresponding PWC-Net variant.}
	\label{fig:flownet_iterres_biocc}	
\end{minipage} \quad 
\begin{minipage}{0.245\textwidth}
	\centering
	\subcaptionbox{Ground-truth occlusion map.\label{fig:occ_compare_a}}{\includegraphics[width=0.98\linewidth]{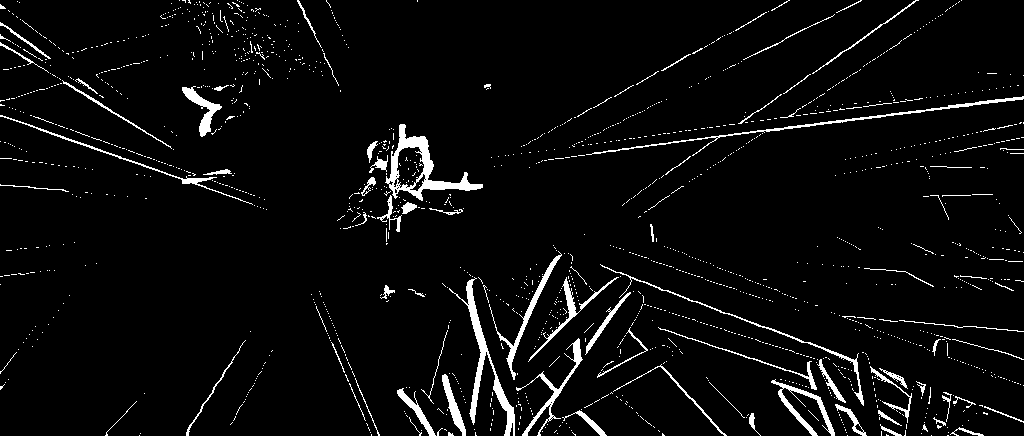}} \\[2mm]
	\subcaptionbox{Reconstructed occlusion map from a quarter resolution of \emph{(\subref{fig:occ_compare_a})}.\label{fig:occ_compare_b}}{\includegraphics[width=0.98\linewidth]{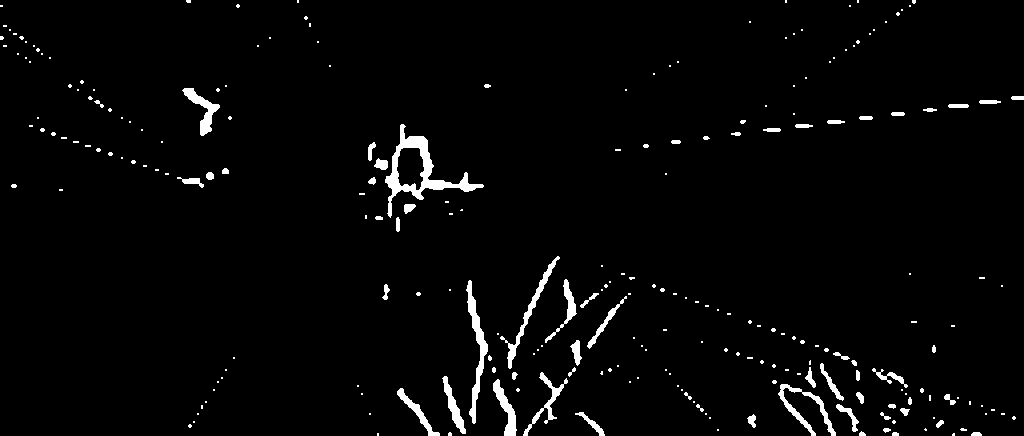}} 
	\smallskip
	\caption{\textbf{Oracle study} showing the limitation of outputting low-resolution occlusion maps.}
	\label{fig:occ_compare}
\end{minipage}
\vspace{-0.5em}
\end{figure*}


Iteratively re-using residual subnetworks and adding occlusion decoders are independent, easily combined together, and adaptable to many types of optical flow base networks.
Beyond simply adopting these two concepts, we additionally propose several ideas to improve the accuracy further in a joint estimation setup: 
\textit{(i)} bi-directional estimation, \textit{(ii)} bilateral refinement of flow and occlusion, and \textit{(iii)} an occlusion upsampling layer.

\myparagraph{Bi-directional estimation.}
Based on the basic IRR setup for joint flow and occlusion estimation in \cref{fig:intro_d}, we first perform bi-directional flow and occlusion estimation by simply switching the order of the input feature maps for the decoder \cite{Meister:2018:ULO,Wang:2018:OAU}.
This yields backward flow $\mv{f}_\text{bw}^i$ and occlusion $\mv{o}^i_2$ in the second frame.
Note that bi-directional estimation requires no extra convolutional weights as it again re-uses the same shared decoders. 
As we shall see, estimating both forward and backward flow together yields at most minor accuracy improvements itself, but we find that exploiting forward-backward consistency is crucial for estimating more accurate occlusions.

\myparagraph{Bilateral refinement of flow and occlusion.}
Blurry estimates, particularly near motion boundaries, have recently been identified as a main limitation of standard optical flow decoders in CNNs.
To address this, bilateral filters or local attention maps \cite{Harley:2017:SAC,Hui:2018:LFN} have been proposed as viable solutions.
We also adopt this idea in our setup, extend it to refine optical flow \emph{and} occlusion using bilateral filters, but with \emph{weight sharing} across all iteration steps.

Similar to Hui \etal~\cite{Hui:2018:LFN}, we construct learned bilateral filters individualized to each pixel and apply them to each flow component $u, v$ and the occlusion separately:
\begin{subequations}
  \begin{align}
    \tilde{f}_{\text{fw}, u}^i(x, y) &= g_\text{fw}(x, y) * f_{\text{fw}, u}^i(x, y) \\
    \tilde{f}_{\text{fw}, v}^i(x, y) &= g_\text{fw}(x, y) * f_{\text{fw}, v}^i(x, y) \\
    \tilde{o}_1^i(x, y) &= g_{o}(x, y) * o_1^i(x, y), 
  \end{align}  
\end{subequations}
where, \eg, $\tilde{f}_{\text{fw}, u}^i(x, y)$ is the filtered horizontal flow at $(x, y)$, $g_\text{fw}(x, y)$ is the $w \times w$ learned bilateral filter kernel for flow at $(x, y)$, and $f_{\text{fw}, u}^i(x, y)$ is the $w \times w$ patch of the horizontal flow centered at $(x, y)$. 
Note that we construct the kernels for flow and occlusion separately as motion and occlusion boundaries are not necessarily aligned.

For constructing the bilateral filter for the flow, we follow the strategy of Hui \etal~\cite{Hui:2018:LFN}, and for occlusion we input occlusion estimates, a feature map, and a warped feature map from the other temporal direction.
One important difference to \cite{Hui:2018:LFN} is that we do not need separate learnable convolutional weights for every iteration step or every pyramid level.
Our IRR design enables re-using the same weights for constructing the bilateral filters for all iteration steps or pyramid levels.
In case of adapting to PWC-Net, our bilateral refinement adds only 0.69M parameters, which is $2.4\times$ less than the scheme of \cite{Hui:2018:LFN}, adding 1.66M parameters.

\myparagraph{Occlusion upsampling layer.} 
One common trait of FlowNet \cite{Dosovitskiy:2015:FLO} and PWC-Net \cite{Sun:2017:PWC} is that the output resolution of flow from the CNN is a quarter ($\frac{1}{4}H \times \frac{1}{4}W$) of the input resolution ($H \times W$), which is then bilinearly upscaled to the input resolution. 
The reasons for not directly estimating at full resolution are a marginal accuracy improvement and the GPU computation and memory overhead.

Yet for estimating occlusion, there is a significant accuracy loss when estimating only at a quarter resolution.
On the Sintel dataset, we conduct an oracle study by downscaling the ground-truth occlusion maps to a quarter size and then upscaling them back.
The F-score of the reconstructed occlusion maps was $0.777$, suggesting a significant accuracy limitation.
As seen in \cref{fig:occ_compare}, quarter-resolution occlusion maps cannot really represent fine occlusions, through which the major loss in F-score occurs.
This strongly emphasizes the importance of estimating at full resolution.

\begin{figure}[t]
\centering     
\includegraphics[width=0.98\linewidth]{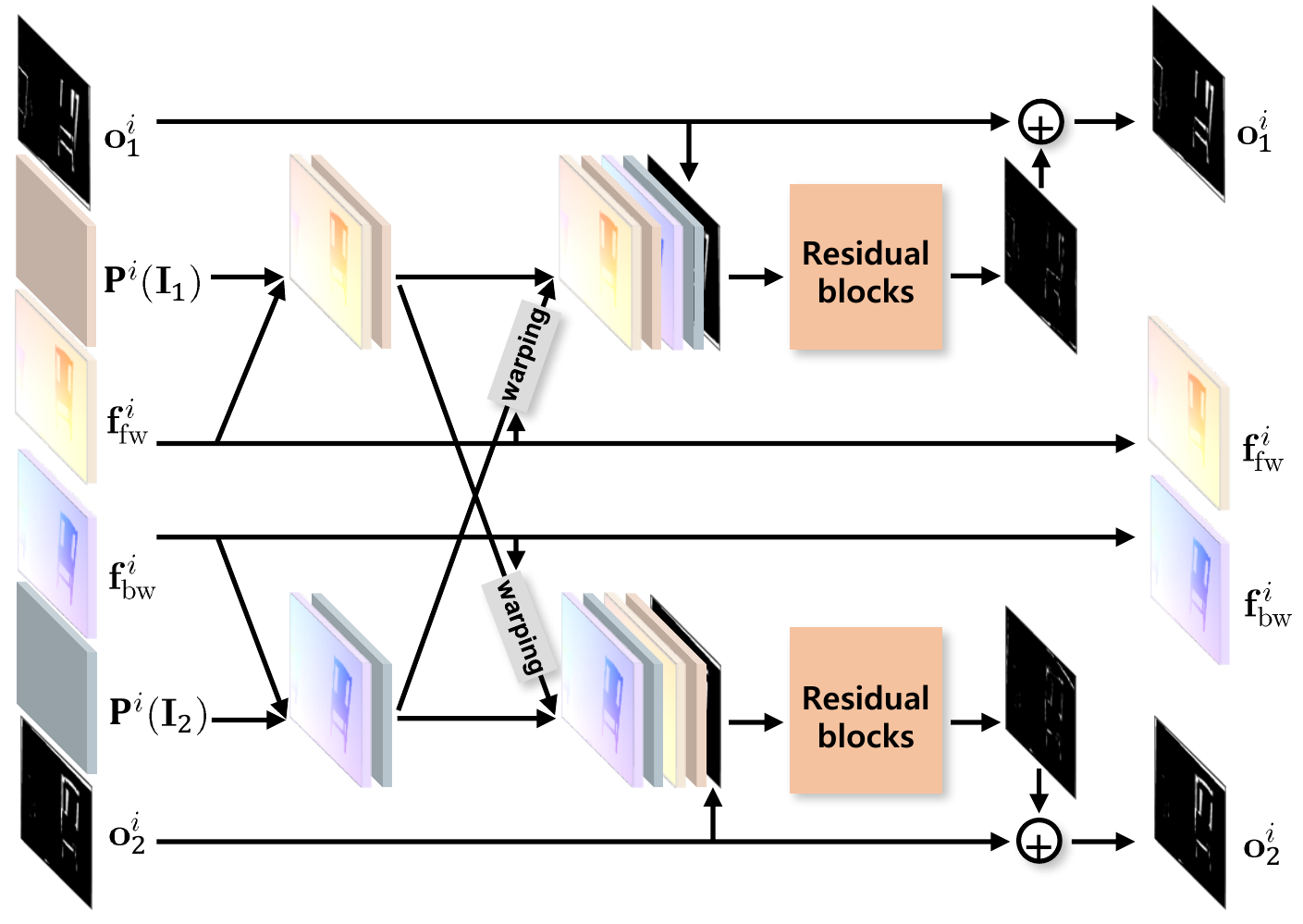}
\caption{\textbf{Occlusion upsampling layer}: Inputs are bi-linearly upsampled flow and upsampled occlusion using nearest neighbor. 
Residual blocks then improve the occlusion accuracy using residual occlusion updates at the full output resolution.}
\label{fig:occ_up}
\vspace{-0.5em}
\end{figure}

To estimate more accurate occlusion at higher resolution, we attach an upsampling layer at the end of network, \cf \cref{fig:flownet_iterres_biocc}, to upscale optical flow and occlusion together back to the input resolution.
\cref{fig:occ_up} illustrates our upsampling layer. 
For optical flow, we found bilinear upsampling to be sufficient. 
For occlusion, we first perform nearest-neighbor upsampling, which is fed into a CNN module to estimate the residual occlusion on the upsampled occlusion map.
The CNN module consists of three residual blocks \cite{Lim:2017:EDR}, which receive flow, a feature map from the encoder, warped flow, and a warped feature map from the other temporal direction.
Putting the warped feature map and flow from the other direction enables exploiting the classical forward-backward consistency for estimating the occlusion.
We provide further details in the supplemental material.

\myparagraph{Initialization.}
To bootstrap our iterative estimation, we input zero as initial optical flow (\ie~$\mv{f}^0_\text{fw}$ and $\mv{f}^0_\text{bw}$) and occlusion (\ie~$\mv{o}^0_1$ and $\mv{o}^0_2$) into the first stage.
Note that 0 indicates non-occluded (visible) and 1 indicates occluded. 

\myparagraph{Training loss.}
Let $N$ be the total number of steps in our iterative setting.
Then we predict a set of forward optical flow maps $\mv{f}^i_\text{fw}$, backward optical flow $\mv{f}^i_\text{bw}$, occlusion maps in the first image $\mv{o}^i_1$ and in the second image $\mv{o}^i_2$ for each iteration step, where $i=1,\ldots,N$.
Forward and backward optical flow are supervised using the $\mathcal{L}_{2,1}$ norm as
\begin{equation}
l^i_\text{flow} = \tfrac{1}{2} \sum \big( \lVert \mv{f}^i_\text{fw} - \mv{f}_{\text{fw},\text{GT}} \rVert_2 + \lVert \mv{f}^i_\text{bw} - \mv{f}_{\text{bw},\text{GT}} \rVert_2 \big),
\label{eq:loss_flow}
\end{equation}
whereas for the supervision of the two occlusion maps we use a weighted binary cross-entropy
\begin{equation}
\begin{split}
l^i_\text{occ} = - \tfrac{1}{2} \sum \big( w^i_1 o^i_1 \log o_{1,\text{GT}} & + \bar{w}^i_1 ( 1\!-\!o^i_1 ) \log (1\!-\!o_{1,\text{GT}})  \\ 
 + w^i_2 o^i_2 \log o_{2,\text{GT}} & + \bar{w}^i_2 ( 1\!-\!o^i_2 ) \log (1\!-\!o_{2,\text{GT}}) 
 \big).
\label{eq:loss_occ}
\end{split}
\end{equation}
Here, we apply the weights $w^i_1 = \frac{H \cdot W}{\sum{o^i_1} + \sum{o_{1,\text{GT}}}}$ and $\bar{w}^i_1 = \frac{H \cdot W}{\sum{(1-o^i_1)} + \sum{(1-o_{1,\text{GT}})}}$ to take into account the number of predictions and true labels.

Our final loss is the weighted sum of the two losses above, taken over all iteration steps using the same multi-scale weights $\alpha_s$ as in the original papers. 
In case of FlowNet \cite{Dosovitskiy:2015:FLO}, the final loss becomes
\begin{equation}
l_\text{FlowNet} = \frac{1}{N}\sum_{i=1}^N \sum_{s=s_0}^S \alpha_s (l^{i,s}_\text{flow} + \lambda \cdot l^{i,s}_\text{occ}), \label{eq:loss_final_flownet}
\end{equation}
where $s$ denotes the scale index given in Fig.~3 of \cite{Dosovitskiy:2015:FLO}. 
In case of PWC-Net \cite{Sun:2017:PWC}, the number of scales is equal to the number of iterations, hence the final loss is
\begin{equation}
l_\text{PWC-Net} = \frac{1}{N}\sum_{i=1}^N \alpha_i (l^{i}_\text{flow} + \lambda \cdot l^{i}_\text{occ}). \label{eq:loss_final_pwc}
\end{equation}
$\lambda$ weighs the flow against the occlusion loss.
In every iteration, we calculate the $\lambda$ that makes the loss of the flow and the occlusion be equal.
We empirically found that this strategy yields better accuracy than just using a fixed trade-off.

\section{Experiments}
\label{sec:Experiment}

\subsection{FlyingChairsOcc dataset}
Lacking a suitable dataset, we create our own dataset for the supervision of bi-directional flow and the two occlusion maps, with ground truth for forward flow, backward flow, and occlusion maps at the first and second frame.
To build the dataset, we follow the exact protocol of the FlyingChairs dataset \cite{Dosovitskiy:2015:FLO}.
We refer to this dataset as \emph{FlyingChairsOcc}.

We crawl $964$ background images with a resolution of $1024 \times 768$ from Flickr and Google using the keywords \emph{cityscape}, \emph{street}, and \emph{mountain}.
As foreground objects, we use $809$ chair images rendered from CAD models with varying views and angles \cite{Aubry:2014:LFN}.
Then we follow the exact protocol of \cite{Dosovitskiy:2015:FLO} for generating image pairs, including the number of foreground objects, object size, and random parameters for generating the motion of each object. 
As the motion is parametrized by a $3 \times 3$ matrix, it is easy to calculate not only backward ground-truth flow but also occlusion maps by conducting visibility checks. 
The number of images in the training and validation sets are the same as in FlyingChairs (\ie $22232$ and $640$, respectively).

\begin{figure*}[t]
\centering     
\includegraphics[width=\textwidth]{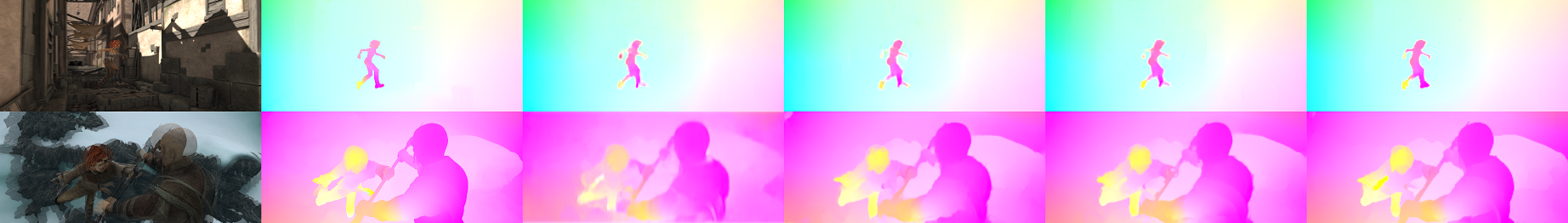}
\caption{\textbf{Qualitative examples from the ablation study on PWC-Net}: \emph{(left to right)} overlapped input images, ground-truth flow, the original PWC-Net \cite{Sun:2017:PWC}, our PWC-Net with IRR, our PWC-Net with Bi-Occ-IRR, and our full model (\ie IRR-PWC).}
\label{fig:pwc_ablation}
\vspace{-0.5em}
\end{figure*}

\subsection{Implementation details}
\paragraph{Training details.}
We follow the training settings of FlowNet respective PWC-Net for a fair comparison.
We use the same geometric and photometric augmentations with additive Gaussian noise as described in \cite{Ilg:2017:FN2}.
After applying the geometric augmentation on the occlusion ground truth, we additionally check for pixels moving outside of the image boundary (\ie~out-of-bound pixels) and set them as occluded.
Note that no multi-stage training is needed.

We first train the proposed model on our FlyingChairs\-Occ dataset with learning rate schedule $S_\text{short}$ (instead of $S_\text{long}$), described in \cite{Ilg:2017:FN2}.
Next, we fine-tune on the Fly\-ing\-Things\-3D-subset dataset \cite{Mayer:2016:ALD}, which contains much larger displacements; we use half the $S_\text{fine}$ learning rate schedule \cite{Ilg:2017:FN2}.
We empirically found that using shorter schedules was enough as our model converged faster.
We finally fine-tune on different public benchmark datasets, including Sintel \cite{Butler:2012:NOS} and KITTI \cite{Geiger:2012:AWR}, following the fine-tuning protocol of \cite{Sun:2018:MMS}.
We use a smaller minibatch size of $4$, as our model implicitly increases the batch size by performing iterative bi-directional estimation with a single model.

Lacking other ground truth, we only use the forward flow and the occlusion map for the first frame for supervision on Sintel; for KITTI we only use the forward flow.
Importantly, our model is still trainable when ground truth is available only for one direction (\eg,~forward flow with occlusion map at the first frame), since both temporal directions share the same ``unidirectional'' decoder. 

\subsection{Ablation study}

To see the effectiveness of each proposed component, we conduct an ablation study by training our model in multiple settings. 
All models are trained on the FlyingChairsOcc dataset with the $S_\text{short}$ schedule and tested on multiple datasets to assess generalization across datasets.
We use a minibatch size of 4 when either bi-directional estimation or iterative residual refinement is on, or the original minibatch size of 8, otherwise.
For a simpler ablation study, we use two iteration steps when applying IRR on FlowNet \cite{Dosovitskiy:2015:FLO}.

\begin{table}[t]
\centering
\scriptsize
\setlength\tabcolsep{4.15pt}
\begin{tabular*}{\columnwidth}{@{}llllllllS[table-format=2.1,retain-explicit-plus]@{}}
	\toprule
	\multirow{2}{*}{  } & \multirow{2}{*}{\rotatebox[origin=c]{90}{Bi}} & \multirow{2}{*}{\rotatebox[origin=c]{90}{Occ}} & \multirow{2}{*}{\rotatebox[origin=c]{90}{IRR}}& {Chairs} & {ChairsOcc} & {Sintel Clean} & {Sintel Final} & \multicolumn{1}{l}{Rel.} \\
      &       &        &     &     {Full}   &      {Validation}  &   {Training}   &   {Training}  & \multicolumn{1}{l}{Param.} \\
	\midrule
	\multirow{9}{*}{{\rotatebox[origin=c]{90}{FlowNet \cite{Dosovitskiy:2015:FLO}}}}  &	&	&		& 2.39	& 2.27			& 4.35			& 5.44  & 0\%	\\
	          &\cmark	&        &          & 2.43	& 2.30			& 4.40 			& 5.53 		& 0\% \\
	          &			& \cmark &          & 2.29	& 2.18 (0.690) 	& 4.26 (0.521)  & 5.51 (0.493) & +38.5\% \\
	          & 		&        & \cmark   & 2.36 	& 2.22			& 3.77  		& 5.00    		& 0\% \\
	          &\cmark	& \cmark &          & 2.31 	& 2.20 (0.691)  & 4.21 (0.515) 	& 5.46 (0.488) & +38.5\% \\
	          &\cmark	&        & \cmark   & 2.14 	& 2.00 			& 3.45 			& 4.96 		& 0\%	\\
	          &			& \cmark & \cmark   & 2.22 	& 2.10 (0.689) 	& 3.56 (0.507) 	& 5.03 (0.486) & +38.5\% \\ 
	          &\cmark	& \cmark & \cmark   & 2.05 	& 1.91 (0.699) 	& 3.40 (0.528) 	& 5.08 (0.502) & +38.5\% \\ 
	          &\cmark	& \cmark & \cmark + & \bfseries 1.92  & \bfseries 1.77 (0.736)  & \bfseries 3.32 (0.596) & \bfseries 4.92 (0.560) & +40.7\% \\ \midrule
	\multirow{9}{*}{{\rotatebox[origin=c]{90}{PWC-Net \cite{Sun:2017:PWC}}}} &	&	&		& 2.03 & 1.89 & 3.13 & 4.41 & 0\% \\
	          &\cmark	&        &			& 2.06 & 1.87 & 2.98 & 4.14   & 0\% \\
	          &			& \cmark &          & 1.94 & 1.79 (0.706) & 3.16 (0.616) & 4.35 (0.581) & +87.4\%\\
	          &			&        & \cmark   & 2.01 & 1.83 & 2.79 & 4.10  & -61.2\% \\
	          &\cmark	& \cmark &          & 1.99 & 1.82 (0.696) & 3.01 (0.618) & 4.39 (0.581) & +87.4\% \\
	          &\cmark	&        & \cmark   & 2.08 & 1.90 & 2.80 & 4.13 & -61.2\% \\
	          &			& \cmark & \cmark   & 1.91 & 1.73 (0.700) & 2.64 (0.630) & 4.09 (0.593) & -34.7\% \\
	          &\cmark	& \cmark & \cmark   & 1.98 & 1.81 (0.698) & 2.69 (0.633) & 4.03 (0.598) & -34.7\%\\ 
	          &\cmark	& \cmark & \cmark + & \bfseries 1.67 & \bfseries 1.48 (0.757) & \bfseries 2.34 (0.677) & \bfseries 3.95 (0.624) & -26.4\% \\ \bottomrule	
\end{tabular*}
\caption{\textbf{Ablation study of our design choices on the two baseline models.} The numbers indicate the average end-point error (EPE) for optical flow (the lower the better) and the average F1-score for occlusion in parentheses, where available (the higher the better). \textbf{Bi}: Bi-directional estimation, \textbf{Occ}: Joint occlusion estimation, \textbf{IRR}: Iterative residual refinement, \textbf{IRR+}: Iterative residual refinement including bilateral refinement and occlusion upsampling layer. The final column reports the relative changes on the number of parameters comparing to the vanilla baseline.}
\label{table:ablation_study}
\vspace{-0.5em}
\end{table}

\cref{table:ablation_study} assesses the optical flow in terms of the average end-point error (EPE) and occlusion estimation with the average F1-score, if applicable for the respective configuration.
In contrast to findings in recent work \cite{Ilg:2018:OMD}, estimating occlusion together yields a gradual improvement of the flow of up to $5\%$ on the training domain, and an even bigger improvement across different datasets when combined on top of bi-directional estimation (Bi) or IRR.
We believe this to mainly stem from using a separate occlusion decoder instead of a joint decoder \cite{Ilg:2018:OMD}.
Bi-directional estimation by itself yields at most a marginal improvement on flow, but it is important for the input of the occlusion upsampling layer, which brings very large benefits on occlusion estimation.
Iterative residual refinement yields consistent improvements in flow accuracy on the training domain, and perhaps surprisingly \emph{a much better generalization across datasets}, with up to 10\% improvement in EPE. 
We presume that this better generalization comes from training a single decoder to handle feature maps from all iteration steps or pyramid levels, which encourages generalization even across datasets.
The benefits of using IRR become even clearer when combined with other components.
For example, FlowNet with Bi, Occ, and IRR demonstrates up to 20\% improvement in EPE on Sintel Clean compared to only using Bi and Occ. 
Additionally, the bilateral refinement and the upsampling layer significantly improve the accuracy of both flow and occlusion with a small overhead of only 0.83M parameters.
For PWC-Net, we obtain a significant accuracy boost of 17.7\% on average over the baseline, while reducing the number of parameters by 26.4\%.
We name the full versions of the models including all modules \textbf{IRR-FlowNet} and \textbf{IRR-PWC}.
\cref{fig:pwc_ablation} highlights the improvement of the flow from our proposed components with qualitative examples.
Please note the completeness and sharp boundaries.

\myparagraph{Bilateral refinement.} 
We compare our bilateral refinement layer with the refinement layer of LiteFlowNet \cite{Hui:2018:LFN} based on a PWC-Net with Bi, Occ, and IRR components enabled. 
\cref{table:bilateral_refine} shows that the benefit of our design choice (\ie sharing weights) holds for bilateral refinement as well, yielding better accuracy for flow and particularly for occlusion, with $2.5\times$ fewer parameters than that of \cite{Hui:2018:LFN}.

\begin{table}[t]
\centering
\scriptsize
\setlength\tabcolsep{3.85pt}
\begin{tabular*}{\columnwidth}{@{}lllllS[table-format=2.1,retain-explicit-plus]@{}}
	\toprule
	\multirow{2}{*}{Method} & {Chairs} & {ChairsOcc} & {Sintel Clean} & {Sintel Final} & \multicolumn{1}{l}{Rel.} \\
          &     {Full}   &      {Validation}  &   {Training}   &   {Training}  & \multicolumn{1}{l}{Param.} \\
    \midrule
    No refinement & 1.98 & 1.81 (0.698) & 2.69 (0.633) & 4.03 (0.598) & 0 \% \\
	Ours & \bfseries 1.66 & \bfseries 1.45 (0.735) & \bfseries 2.32 (0.648) & 3.90 {\bfseries (0.602)} & +12.3 \% \\ 
	LiteFlowNet's \cite{Hui:2018:LFN} & 1.74 & 1.58 (0.688) & 2.34 (0.596) & {\bfseries 3.86} (0.543) & +29.5 \% \\ \bottomrule
	
\end{tabular*}
\caption{Comparison of our bilateral refinement layer against that of LiteFlowNet \cite{Hui:2018:LFN}.}
\label{table:bilateral_refine}
\vspace{-0.5em}
\end{table}

{
\begin{table}[t]
\centering
\scriptsize
\setlength\tabcolsep{4.85pt}
\begin{tabular*}{\columnwidth}{@{}lllllS[table-format=1.2,retain-explicit-plus]@{}}
	\toprule
	\multirow{2}{*}{Method} & {Chairs} & {ChairsOcc} & {Sintel Clean} & {Sintel Final} & \multicolumn{1}{l}{Rel.} \\
          &     {Full}   &      {Validation}  &   {Training}   &   {Training}  & \multicolumn{1}{l}{Param.} \\
    \midrule
    No upsampling & \bfseries 1.66 & {\bfseries 1.45} (0.735) & {\bfseries 2.32} (0.648) & {\bfseries 3.90} (0.602) & 0 \% \\
	Ours 	 & 1.67 & 1.48 {\bfseries (0.757)} & 2.34 {\bfseries (0.677)} & 3.95 {\bfseries (0.624)} & +0.49 \% \\ 
	\cite{Ilg:2017:FN2,Ilg:2018:OMD} & 2.18 & 2.01 (0.712) & 2.90 (0.624) & 4.37 (0.577) & +9.21 \% \\ \bottomrule	
\end{tabular*}
\caption{Comparison of our occlusion upsampling layer and the refinement network from FlowNet2 \cite{Ilg:2017:FN2,Ilg:2018:OMD}.}
\label{table:upsampling}
\vspace{-0.5em}
\end{table}
}

\myparagraph{Occlusion upsampling layer.}
Similar to our upsampling layer, \cite{Ilg:2018:OMD} uses a refinement network from FlowNet2 \cite{Ilg:2017:FN2} to upsample the intermediate quarter-resolution outcome back to the original resolution.
We compare our upsampling layer with the refinement network from \cite{Ilg:2017:FN2,Ilg:2018:OMD}, adding it to our network based on a PWC-Net backbone with Bi, Occ, IRR, and the bilateral refinement layer enabled.
\cref{table:upsampling} shows the clear benefits of using our upsampling layer, yielding significant gains in both tasks while requiring fewer parameters.
The refinement network from FlowNet2 \cite{Ilg:2017:FN2} actually degrades the accuracy of flow estimation.
We presume this may stem from differences in training.
FlowNet2's refinement layer may require piece-wise training, while our model is trained all at once.

\myparagraph{Different IRR steps on FlowNet.} For FlowNet, we can freely choose the number of IRR steps as we iteratively refine previous estimates by re-using a single network.
We try different numbers of IRR steps on vanilla FlowNetS \cite{Dosovitskiy:2015:FLO} (\ie without Bi or Occ) and compare with stacking multiple FlowNetS networks.
All networks are trained on FlyingChairsOcc with the $S_{short}$ schedule, minibatch size of $8$, and tested on Sintel Clean. 
{
\begin{table}[t]
\centering
\scriptsize
\setlength\tabcolsep{4pt}
\begin{tabularx}{\columnwidth}{@{}Xccccc@{}}
    \toprule
    Number of iterations or stacking stages & 1 & 2 & 3 & 4 & 5 \\ \midrule 
    IRR on a single FlowNetS  & \bfseries 4.358 & \bfseries  3.545 & \bfseries 3.325 & \bfseries 3.303 & \bfseries 3.302 \\     
    Stacking multiple FlowNetS & 4.445 & 3.553 & 3.377 & 3.391 & 3.517 \\ \bottomrule    
\end{tabularx}
\caption{$n\times$ IRR \vs~$n\times$ stacking: EPE on Sintel Clean.}
\label{table:iteration_study}
\vspace{-0.5em}
\end{table}
}
As shown in \cref{table:iteration_study}, the accuracy keeps improving with more IRR steps and stably settles at more than $4$ steps.
In contrast, stacking multiple FlowNetS networks overfits on the training data after $3$ steps, and is consistently outperformed by IRR with the same number of stages. 
This clearly demonstrates the advantage of our IRR scheme over stacking: \emph{better accuracy without linearly increasing the number of parameters}.

{
\begin{table}[t]
\centering
\scriptsize
\setlength\tabcolsep{4pt}
\begin{tabular*}{\columnwidth}{@{}l@{\extracolsep{\stretch{1}}}ccccS[table-format=3.2]@{}}
	\toprule
	\multirow{2}{*}{Method} & \multicolumn{2}{c}{Training} & \multicolumn{2}{c}{Test} & {\multirow{2}{*}{Parameters}} \\
	\cmidrule(lr){2-3} \cmidrule(lr){4-5} 
     & Clean & Final & Clean & Final & \\ 
	\midrule
	ContinualFlow\_ROB$^{\dagger\S}$ \cite{Neoral:2018:COO} & -- & -- &3.34& \textbf{4.53}& 14.6 \si{\mega}	\\ 
	MFF$^{\S}$	\cite{Ren:2018:FAM}					  & -- & -- &3.42&4.57& {N/A} \\ 
	\textbf{IRR-PWC (Ours)}					  & (1.92) & (2.51) &3.84 & 4.58 & 6.36 \si{\mega}	\\ 
	PWC-Net+$^{\dagger}$ \cite{Sun:2018:MMS}			  &(1.71)&(2.34)&3.45&4.60& 8.75 \si{\mega} \\ 
	ProFlow$^{\S}$\cite{Maurer:2018:PFL}			  & -- & -- &2.82&5.02& {--} \\ 
	PWC-Net-ft-final \cite{Sun:2018:MMS} 	  &(2.02)&(2.08)&4.39&5.04&	8.75 \si{\mega} \\ 
	DCFlow \cite{Xu:2017:AOF}				&--&--&3.54&5.12&	{--} \\ 
	FlowFieldsCNN \cite{Bailer:2017:CPM}   &--&--&3.78&5.36& 5.00 \si{\mega} \\ 
	MR-Flow \cite{Wulff:2017:OFM}		&1.83&3.59&\textbf{2.53}&5.38& {--} \\ 
	LiteFlowNet \cite{Hui:2018:LFN}		  & (1.35) & (1.78) &4.54&5.38&	5.37 \si{\mega} \\ 
	S2F-IF \cite{Yang:2017:S2F}			& -- & -- &3.50&5.42&	{--} \\  
	SfM-PM \cite{Maurer:2018:SMA}		& -- & --&2.91&5.47& {--} \\  
	FlowFields++ \cite{Schuster:2018:FFA}	&-- & -- &2.94&5.49& {--} \\
	FlowNet2 \cite{Ilg:2017:FN2}	&(2.02) & (3.14) &3.96&6.02& 162.5 \si{\mega} \\  \bottomrule	
\end{tabular*}
\caption{\textbf{MPI Sintel Flow}: Average end-point error (EPE) and number of CNN parameters. $^{\S}$using more than 2 frames, $^{\dagger}$using additional datasets (KITTI and HD1k) for better accuracy.}
\label{table:flow_Sintel}
\vspace{-0.5em}
\end{table}
}

\subsection{Optical flow benchmarks}

We test the accuracy of our IRR-PWC on the public Sintel \cite{Butler:2012:NOS} and KITTI \cite{Geiger:2012:AWR,Menze:2015:OSF} benchmarks.
When fine-tuning, we use the robust training loss as in \cite{Hui:2018:LFN,Sun:2017:PWC,Sun:2018:MMS} for flow, and standard binary cross-entropy for occlusion.
On Sintel Final, our IRR-PWC achieves a new state of the art among 2-frame methods.
Comparing to the PWC-Net baseline (\ie PWC-Net-ft-final) trained in the identical setting, our contributions improve the flow accuracy by $9.18\%$ on Final and $12.36\%$ on Clean, while using $26.4\%$ fewer parameters. 
On KITTI 2015, our IRR-PWC again outperforms all published 2-frame methods, improving over the baseline PWC-Net.

When fine-tuning on benchmarks, our important observations are that our model \emph{(i)} converges much faster than the baseline and \emph{(ii)} overfits to the training split less, demonstrating much better accuracy on the test set despite slightly higher error on training split.
This highlights the benefit of our IRR scheme: better generalization \emph{even on the training domain} as well as across datasets.

{
\begin{table}[t]
\centering
\scriptsize
\setlength\tabcolsep{10pt}
\begin{tabularx}{\columnwidth}{@{}XccS[table-format=2.2,table-space-text-post = \si{\%}]@{}}
	\toprule
	\multirow{2}{*}{Method}  &  \multicolumn{2}{c}{Training} & {Test}  \\ \cmidrule(lr){2-3} \cmidrule(l){4-4} 
     					    			 	& AEPE 	& Fl-all & {Fl-All} \\ \midrule
	MFF$^{\S}$	\cite{Ren:2018:FAM}					&  -- & -- & \bfseries 7.17\% \\
	\textbf{IRR-PWC (Ours)}			 		& (1.63) & (5.32\%) & 7.65\% \\
	PWC-Net+ \cite{Sun:2018:MMS}				& (1.45) & (7.59\%) & 7.72\% \\
	LiteFlowNet \cite{Hui:2018:LFN}			& (1.62) & (5.58\%) & 9.38\% \\
	PWC-Net \cite{Sun:2017:PWC}				& (2.16) & (9.80\%) & 9.60\% \\
	ContinualFlow\_ROB$^{\dagger\S}$ \cite{Neoral:2018:COO} & -- & -- & 10.03\% \\
	MirrorFlow \cite{Hur:2017:MFE}			& -- & 9.98\% & 10.29\% \\
	FlowNet2 \cite{Ilg:2017:FN2}			& (2.30) & (8.61\%) & 10.41\% \\ \bottomrule
\end{tabularx}
\caption{\textbf{KITTI Optical Flow 2015}: Average end-point error (EPE) and outlier rates (Fl-Noc and Fl-all).}
\label{table:flow_KITTI}
\vspace{-0.5em}
\end{table}
}


\subsection{Occlusion estimation}

{
\begin{table}[t]
\centering
\scriptsize
\setlength\tabcolsep{10pt}
\begin{tabularx}{\columnwidth}{@{}XcS[table-format=1.3]S[table-format=1.3]@{}}
	\toprule
	\multirow{2}{*}{Method} & \multirow{2}{*}{Type} & \multicolumn{2}{c}{Sintel Training} \\ \cmidrule(lr){3-4} 
			                 & & {Clean} & {Final} \\ \midrule 
	\textbf{IRR-PWC (Ours)} & supervised & \bfseries 0.712 & \bfseries 0.669 \\
	FlowNet-CSSR \cite{Ilg:2018:OMD}	& supervised & 0.703 & 0.654	\\
	OccAwareFlow \cite{Wang:2018:OAU} & unsupervised & 0.54& 0.48 \\
	Back2FutureFlow \cite{Janai:2018:ULM} & unsupervised & 0.49 & 0.44 \\  
	MirrorFlow \cite{Hur:2017:MFE} & estimated & 0.390 & {--} \\\bottomrule	
\end{tabularx}
\caption{Occlusion estimation results on Sintel Training.}
\label{table:occ_sintel}
\vspace{-0.5em}
\end{table}
}

We finally evaluate the accuracy of occlusion estimation on the Sintel training set as no public benchmarks are available for the task. 
\cref{table:occ_sintel} shows the comparison with state-of-the-art algorithms.
Supervised methods are trained on FlyingChairs and FlyingThings3D; unsupervised methods are trained on Sintel without the use of ground truth. 
We achieve state-of-the-art accuracy with far fewer parameters ($6.00 \si{\mega}$ instead of $110 \si{\mega}$) and much simpler training schedules than the previous state of the art \cite{Ilg:2018:OMD}.


\section{Conclusion}
\label{sec:Conclusion}
We proposed an iterative residual refinement (IRR) scheme based on weight sharing for generic optical flow networks, with additional components for bi-directional estimation and occlusion estimation.
Applying our scheme on top of two representative flow networks, FlowNet and PWC-Net, significantly improves flow accuracy with a better generalization while even reducing the number of parameters in case of PWC-Net. 
We also show that our design choice of jointly estimating occlusion together with flow brings accuracy improvements on both domains, setting the state of the art on public benchmark datasets.
We believe that our powerful IRR scheme can be combined with other baseline networks and can form the basis of other follow-up approaches, including multi-frame methods.

{\small
\balance
\bibliographystyle{ieee_fullname}

}

\normalsize
{
\twocolumn[{
   \newpage
   \null
   \vskip .375in
   \begin{center}
      {\Large \bf \supptitle \par}
      \vspace*{24pt}
      {
      \large
      \lineskip .5em
      \begin{tabular}[t]{c}
         \ifcvprfinal\suppauthor\else Anonymous CVPR submission\\
         \vspace*{1pt}\\
Paper ID \cvprPaperID \fi
      \end{tabular}
      \par
      }
      \vskip .5em
      \vspace*{12pt}
   \end{center}
   }]
}

\appendix
\nobalance

Here, we provide additional details on IRR-PWC, our occlusion upsampling layer, more qualitative examples on the ablation study, as well as a qualitative comparison with the state of the art. 

\section{IRR-PWC}
\label{sec:irr_pwc_figure}

\begin{figure}[t]
\centering
\begin{subfigure}{\linewidth}
  \includegraphics[width=\linewidth]{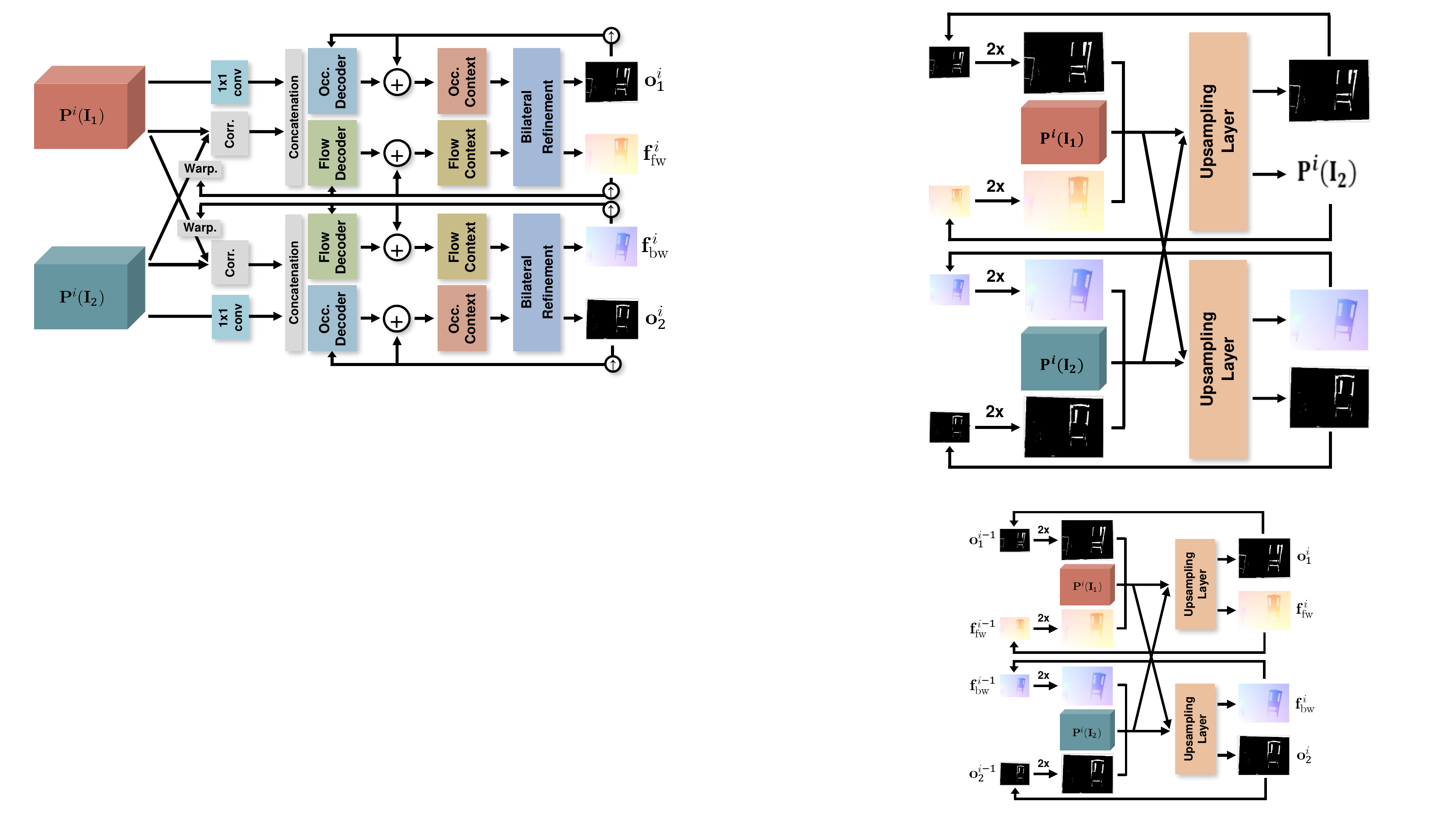}
  \caption{Jointly estimating optical flow and occlusion up to a quarter resolution of the original input, \ie pyramid levels $1 \leq i \leq 5$.}
  \label{fig:irr_pwc_estimator}
\end{subfigure}\\[2mm]
\begin{subfigure}{\linewidth}
  \centering
  \includegraphics[width=0.7\linewidth]{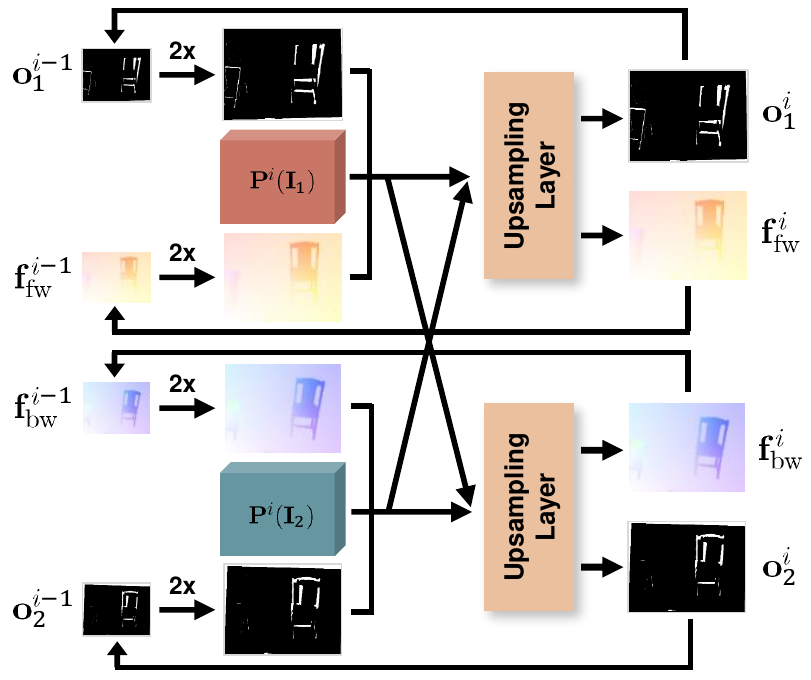}
  \caption{Upsampling optical flow and occlusion using the upsampling layer, \ie pyramid levels $6 \leq i \leq 7$.}
  \label{fig:irr_pwc_upsample}
\end{subfigure}
\caption{\textbf{IRR-PWC}: Our PWC-Net variant with joint optical flow and occlusion estimation based on bi-directional estimation, bilateral refinement, and the occlusion upsampling layer. \emph{(\subref{fig:irr_pwc_estimator})} Our IRR-PWC model jointly estimates optical flow and occlusion up to a quarter resolution of the input image (\ie up to the \nth{5} level), the same as the original PWC-Net. \emph{(\subref{fig:irr_pwc_upsample})} Then, we use our occlusion upsampling layer to upscale the outputs back to the original resolution while improving accuracy.}
\label{fig:irr_pwc_all}
\end{figure}

\cref{fig:irr_pwc_all} shows our IRR-PWC model that jointly estimates optical flow and occlusion using bi-directional estimation, bilateral refinement, and the occlusion upsampling layer.
Given a 7-level feature pyramid as in the original PWC-Net \cite{Sun:2017:PWC}, our IRR-PWC first iteratively and residually estimates optical flow and occlusion up to a quarter resolution of the input image, as shown in \cref{fig:irr_pwc_estimator}.
Then, given the estimates at the \nth{5} level, we show how we use our occlusion upsampling layer in \cref{fig:irr_pwc_upsample} to scale the estimates up to the original resolution.
The upsampling layer upscales the resolution by $2 \times$ at once, and applying the upsampling layer at the \nth{6} and \nth{7} level scales the quarter resolution estimate back to the original resolution.

\cref{fig:occ_up} in the main paper shows the detailed structure of the upsampling layer.
In the following, we describe the details on the residual blocks in the upsampling layer.

\section{Details on the Occlusion Upsampling Layer}
\label{sec:occ_upsampling_layer}

\begin{figure}[t]
\centering
\begin{subfigure}{\linewidth}
  \includegraphics[width=\linewidth]{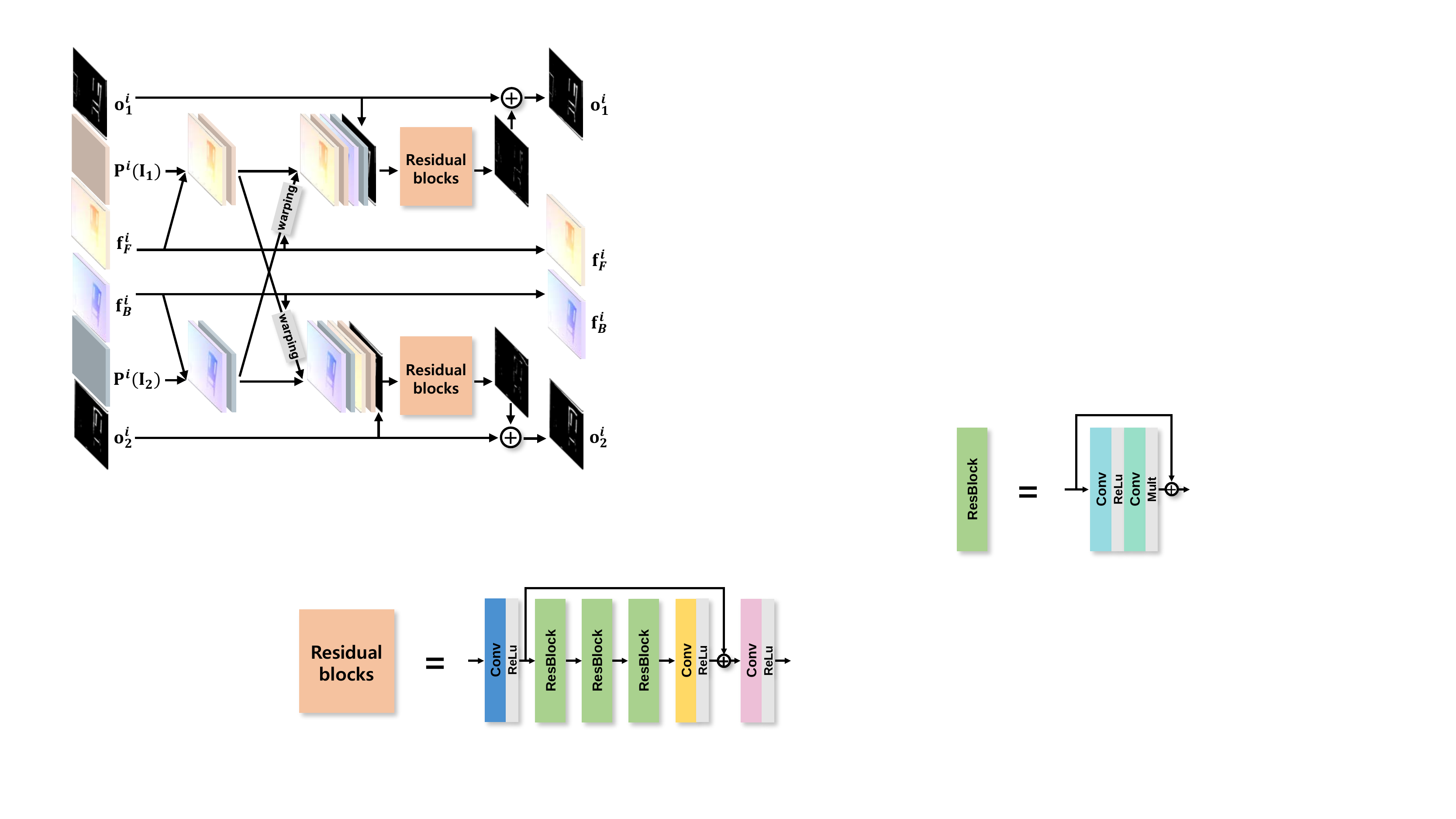}
  \caption{Residual blocks subnetwork.}
  \label{fig:residual_blocks}  
\end{subfigure}\\[1mm]
\begin{subfigure}{\linewidth}
  \centering
  \includegraphics[width=0.5\linewidth]{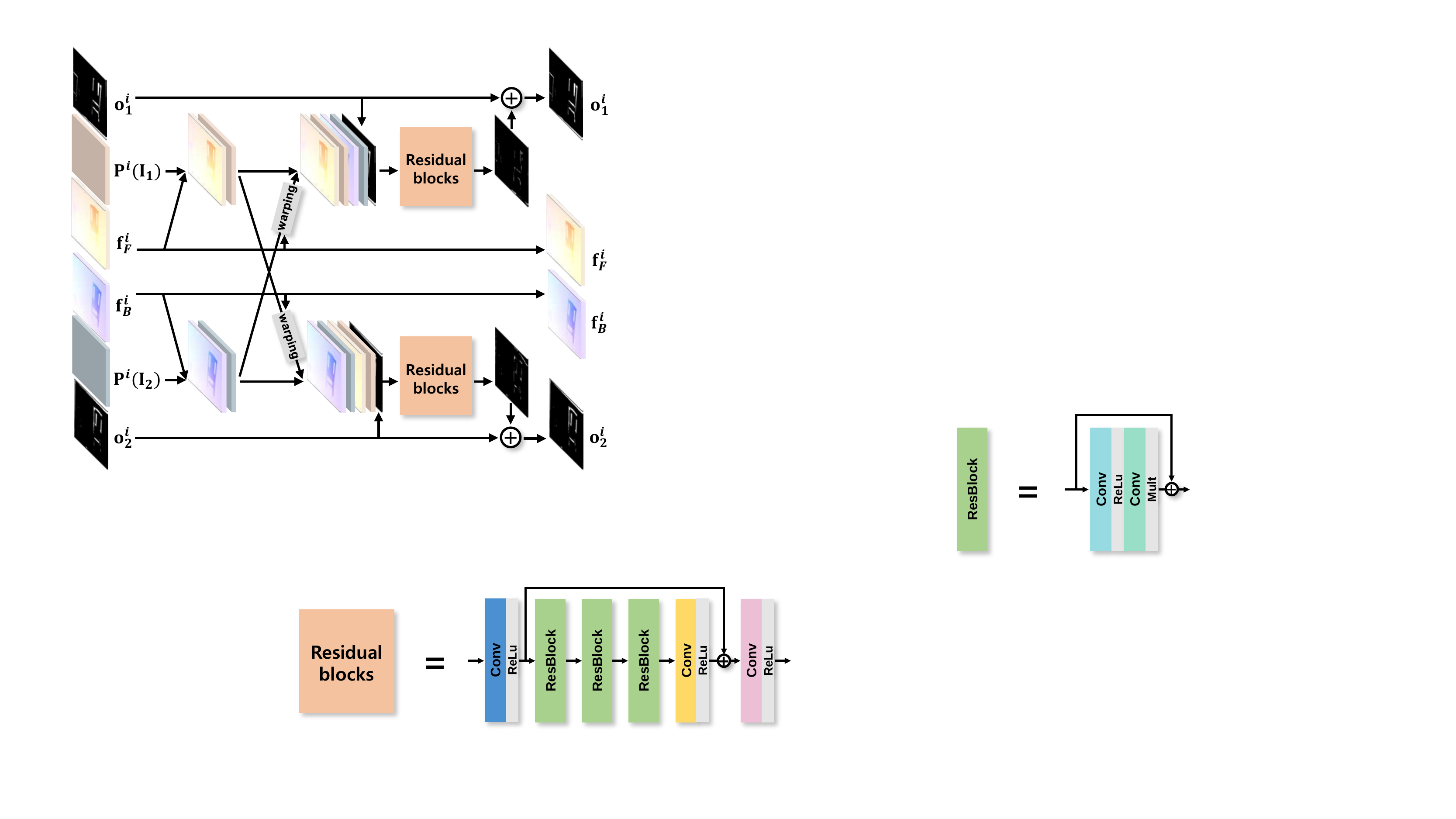}
  \caption{One residual block.}
  \label{fig:one_residual_block}
\end{subfigure}
\caption{\textbf{Residual blocks in the upsampling layer}: \emph{(\subref{fig:residual_blocks})} The \emph{residual blocks} consist of $3$ weight-shared \emph{ResBlock}s with $3$ convolution layers. \emph{(\subref{fig:one_residual_block})} One \emph{ResBlock} consists of \emph{Conv+ReLu+Conv+Mult} operations \cite{Lim:2017:EDR}. }
\label{fig:residual_blocks_all}
\vspace{-0.5em}
\end{figure}

In the occlusion upsampling layer shown in \cref{fig:occ_up} in the main paper, the \emph{residual blocks} \cite{Lim:2017:EDR} are fed a set of feature maps as input and output residual occlusion estimates to refine the upscaled occlusion map from the previous level.
\cref{fig:residual_blocks_all} shows the details of the \emph{residual blocks}. 
As shown in \cref{fig:residual_blocks}, the subnetwork consists of $3$ residual blocks (\ie $3$ \emph{ResBlock}s) with $3$ convolution layers. 
One \emph{ResBlock} consists of \emph{Conv+ReLu+Conv+Mult} operations as shown in \cref{fig:one_residual_block}, \cf \cite{Lim:2017:EDR}.
This sequence of $3$ \emph{ResBlock}s with one convolution layer afterwards estimates the residuals over one convolution output of the input feature maps, and the final convolution layer of the \emph{residual blocks} outputs the residual occlusion.
The number of channels for all convolution layers here is $32$, except for the final convolution layer, which has only $1$ channel for the occlusion output. 

We use weight sharing also on the upsampling layers between bi-directional estimations and between pyramid levels or iteration steps. 
Furthermore, the \emph{ResBlock}s in \cref{fig:residual_blocks} also share their weights, which is different from \cite{Lim:2017:EDR}, where they are not shared.
With this efficient weight-sharing scheme, the occlusion upsampling layer improves the occlusion accuracy by $2.99 \%$ on the training domain (\ie the FlyingChairsOcc dataset) and $4.08 \%$ across datasets (\ie Sintel) with only adding 0.031 \si{\mega} parameters.

{
\begin{figure*}[!b]
\centering
\footnotesize
\setlength\tabcolsep{0.3pt}
\renewcommand{\arraystretch}{0.2}
\begin{tabular}{>{\centering\arraybackslash}m{.25\textwidth} >{\centering\arraybackslash}m{.25\textwidth} >{\centering\arraybackslash}m{.25\textwidth} >{\centering\arraybackslash}m{.25\textwidth}}
	
	\begin{tikzpicture}
    \node[inner sep=0] (img) {\includegraphics[width=\linewidth]{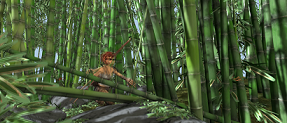}};    
    \node[anchor=north west] at (img.north west){\circled{a}};
	\end{tikzpicture}&
	\begin{tikzpicture}
    \node[inner sep=0] (img) {\includegraphics[width=\linewidth]{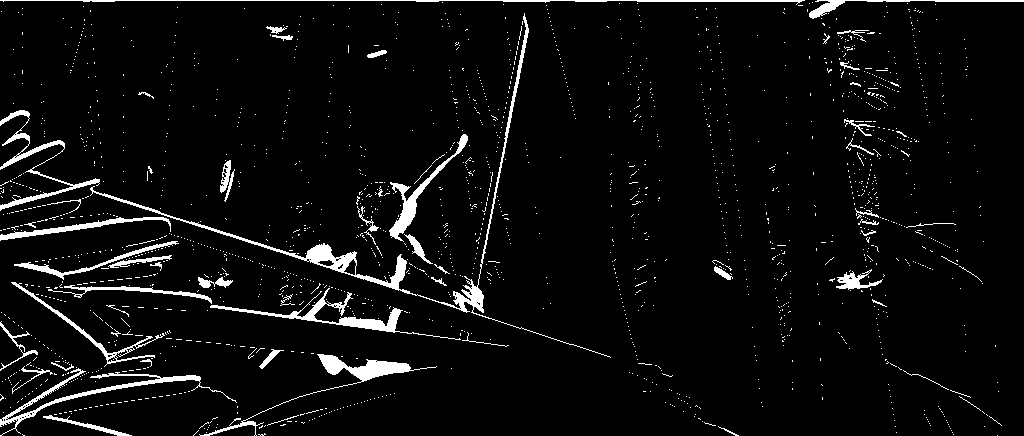}};    
    \node[anchor=north west] at (img.north west){\circled{b}};
	\end{tikzpicture}&
	\begin{tikzpicture}
    \node[inner sep=0] (img) {\includegraphics[width=\linewidth]{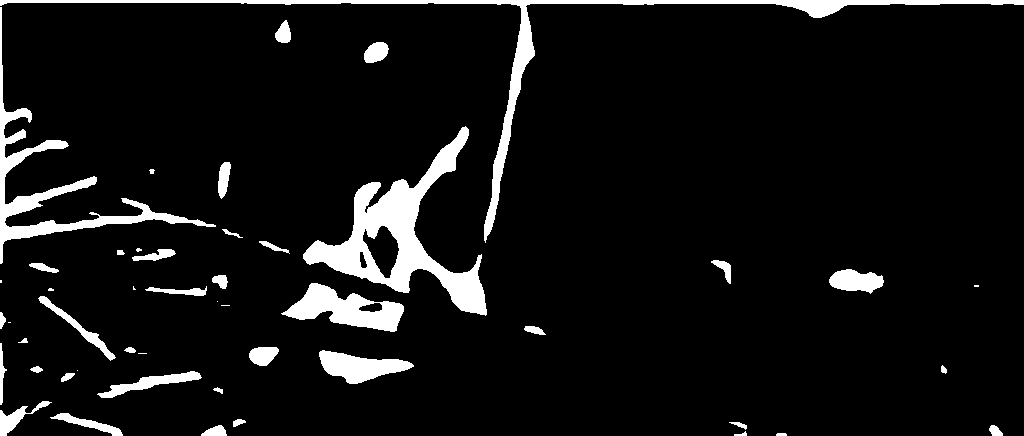}};    
    \node[anchor=north west] at (img.north west){\circled{c}};
	\end{tikzpicture}&
	\begin{tikzpicture}
    \node[inner sep=0] (img) {\includegraphics[width=\linewidth]{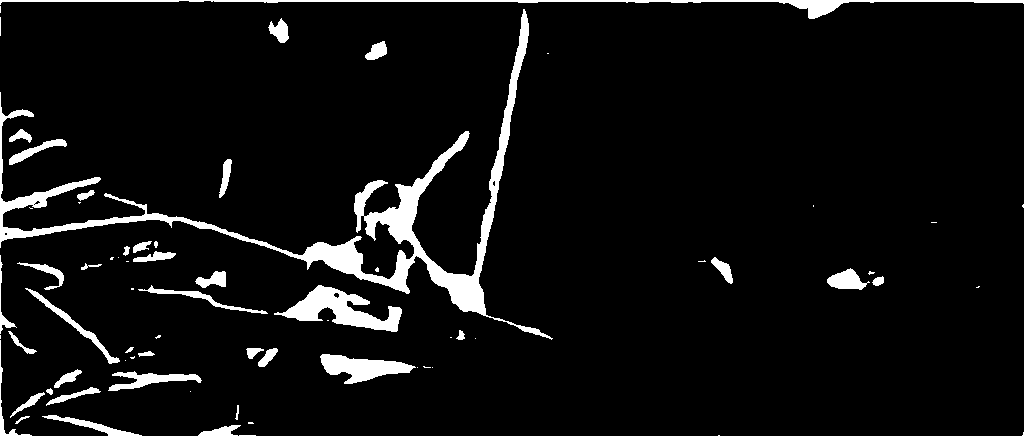}};    
    \node[anchor=north west] at (img.north west){\circled{d}};
	\end{tikzpicture} \\ \\
	
	\begin{tikzpicture}
    \node[inner sep=0] (img) {\includegraphics[width=\linewidth]{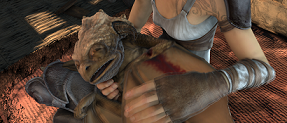}};    
    \node[anchor=north west] at (img.north west){\circled{a}};
	\end{tikzpicture}&
	\begin{tikzpicture}
    \node[inner sep=0] (img) {\includegraphics[width=\linewidth]{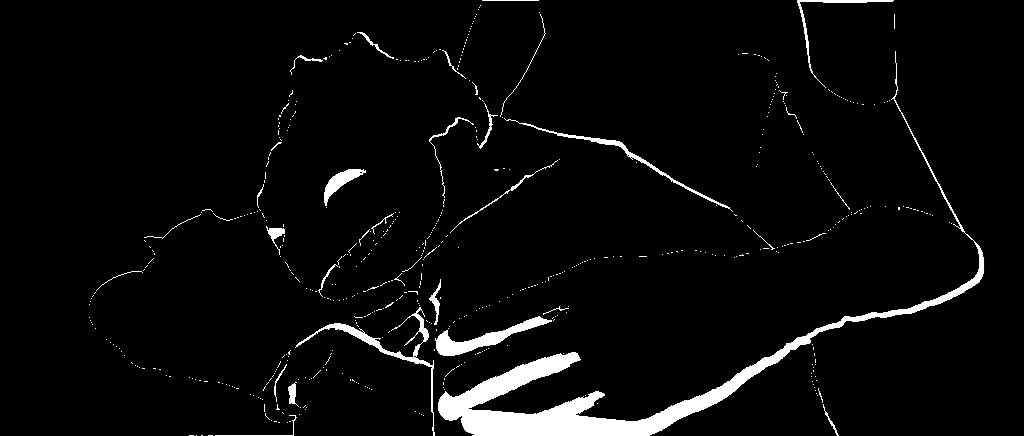}};    
    \node[anchor=north west] at (img.north west){\circled{b}};
	\end{tikzpicture}&
	\begin{tikzpicture}
    \node[inner sep=0] (img) {\includegraphics[width=\linewidth]{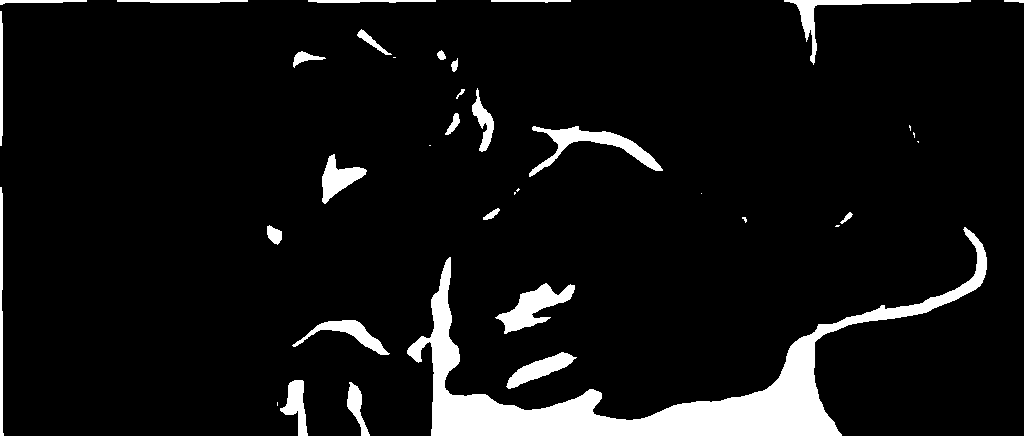}};    
    \node[anchor=north west] at (img.north west){\circled{c}};
	\end{tikzpicture}&
	\begin{tikzpicture}
    \node[inner sep=0] (img) {\includegraphics[width=\linewidth]{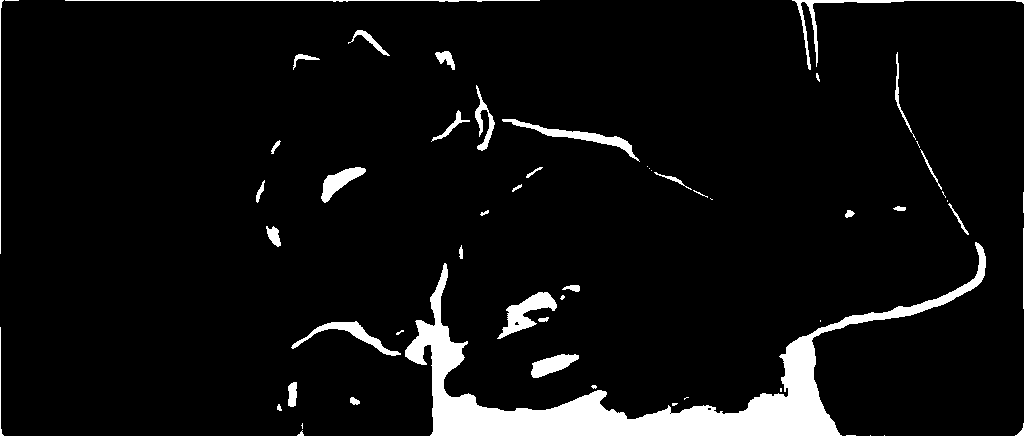}};    
    \node[anchor=north west] at (img.north west){\circled{d}};
	\end{tikzpicture} \\ \\
	
	\begin{tikzpicture}
    \node[inner sep=0] (img) {\includegraphics[width=\linewidth]{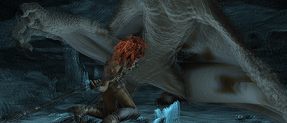}};    
    \node[anchor=north west] at (img.north west){\circled{a}};
	\end{tikzpicture}&
	\begin{tikzpicture}
    \node[inner sep=0] (img) {\includegraphics[width=\linewidth]{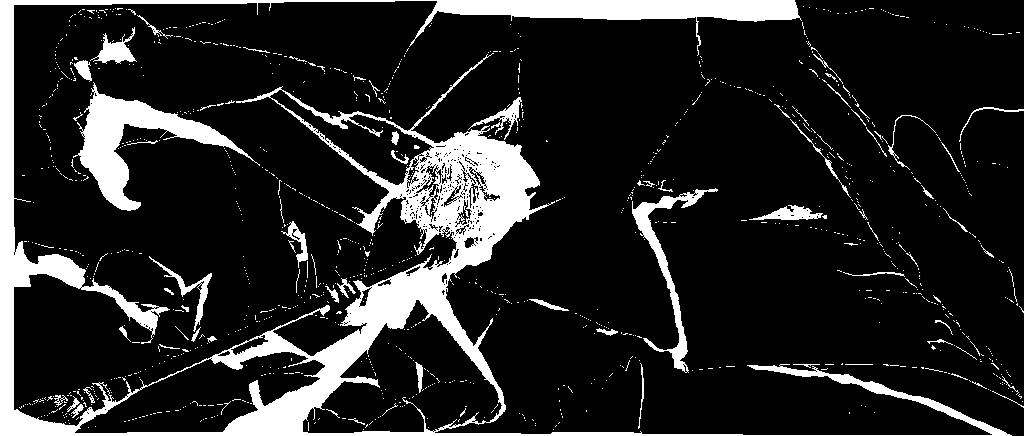}};    
    \node[anchor=north west] at (img.north west){\circled{b}};
	\end{tikzpicture}&
	\begin{tikzpicture}
    \node[inner sep=0] (img) {\includegraphics[width=\linewidth]{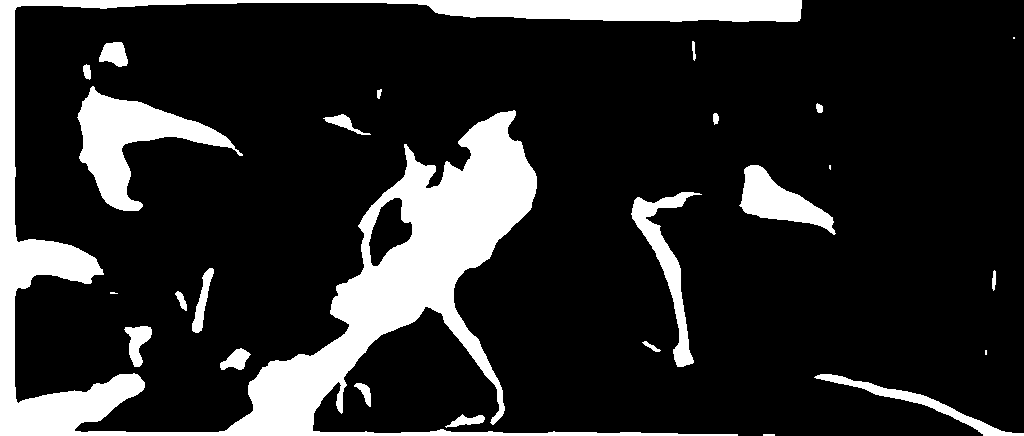}};    
    \node[anchor=north west] at (img.north west){\circled{c}};
	\end{tikzpicture}&
	\begin{tikzpicture}
    \node[inner sep=0] (img) {\includegraphics[width=\linewidth]{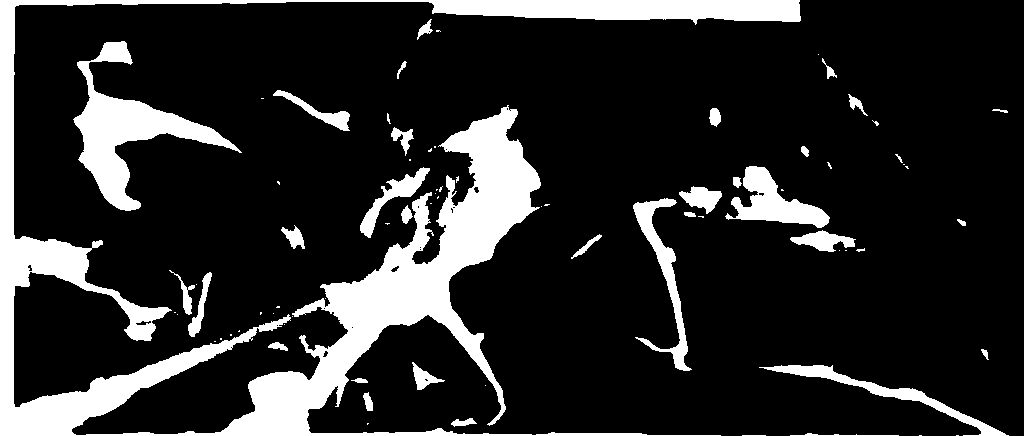}};    
    \node[anchor=north west] at (img.north west){\circled{d}};
	\end{tikzpicture} \\ \\
	
	\begin{tikzpicture}
    \node[inner sep=0] (img) {\includegraphics[width=\linewidth]{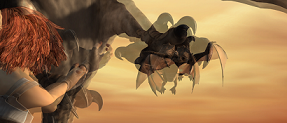}};    
    \node[anchor=north west] at (img.north west){\circled{a}};
	\end{tikzpicture}&
	\begin{tikzpicture}
    \node[inner sep=0] (img) {\includegraphics[width=\linewidth]{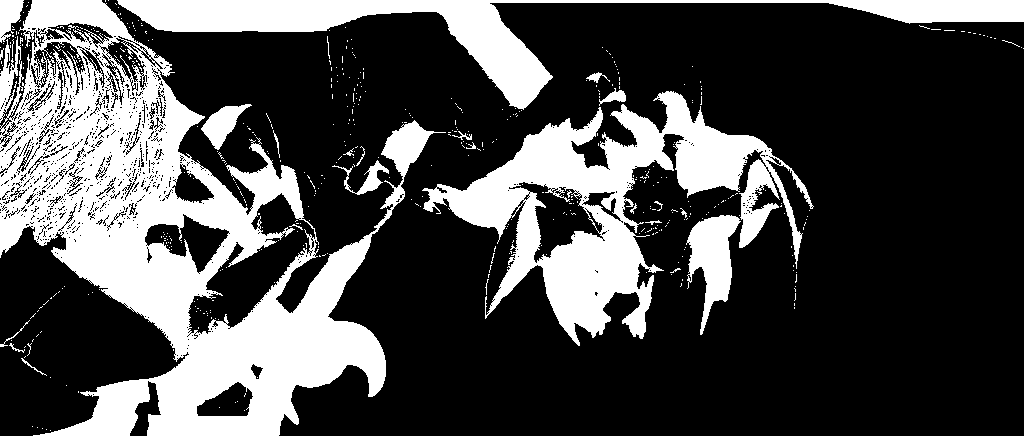}};    
    \node[anchor=north west] at (img.north west){\circled{b}};
	\end{tikzpicture}&
	\begin{tikzpicture}
    \node[inner sep=0] (img) {\includegraphics[width=\linewidth]{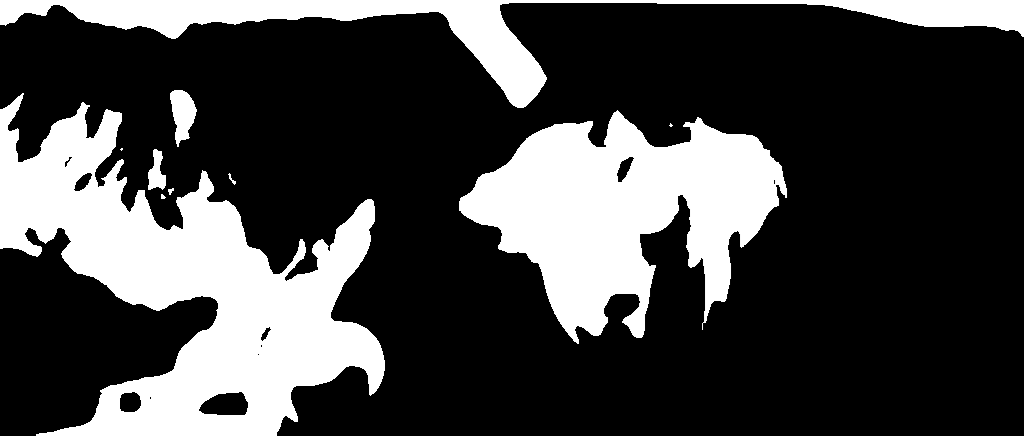}};    
    \node[anchor=north west] at (img.north west){\circled{c}};
	\end{tikzpicture}&
	\begin{tikzpicture}
    \node[inner sep=0] (img) {\includegraphics[width=\linewidth]{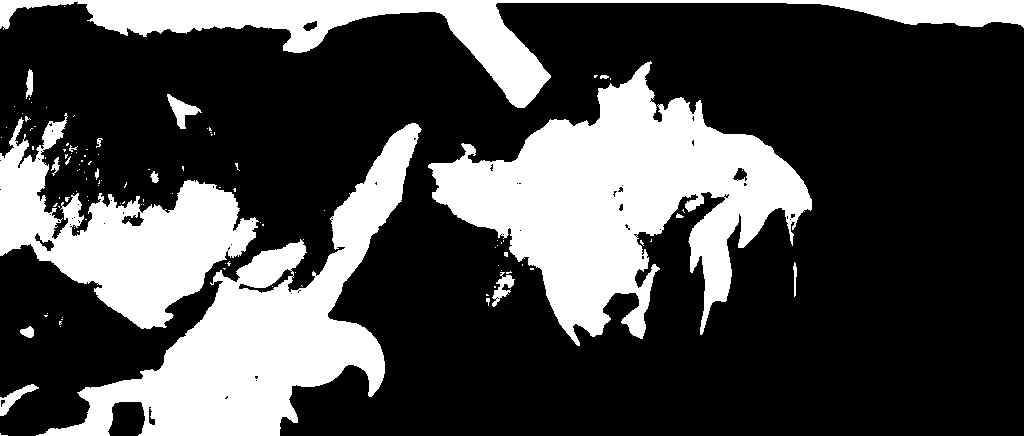}};    
    \node[anchor=north west] at (img.north west){\circled{d}};
	\end{tikzpicture} \\ \\
	
\end{tabular}
\caption {\color{black} \textbf{Qualitative examples of using the occlusion upsampling layer}: \emph{(a)} overlapped input images, \emph{(b)} occlusion ground truth, \emph{(c)} without using the occlusion upsampling layer, and \emph{(d)} with using the occlusion upsampling layer. The occlusion upsampling layer makes occlusion estimates much sharper along motion boundaries and detects additional thinly-shaped occlusions.}
\label{fig:occ_ablation}
\end{figure*}
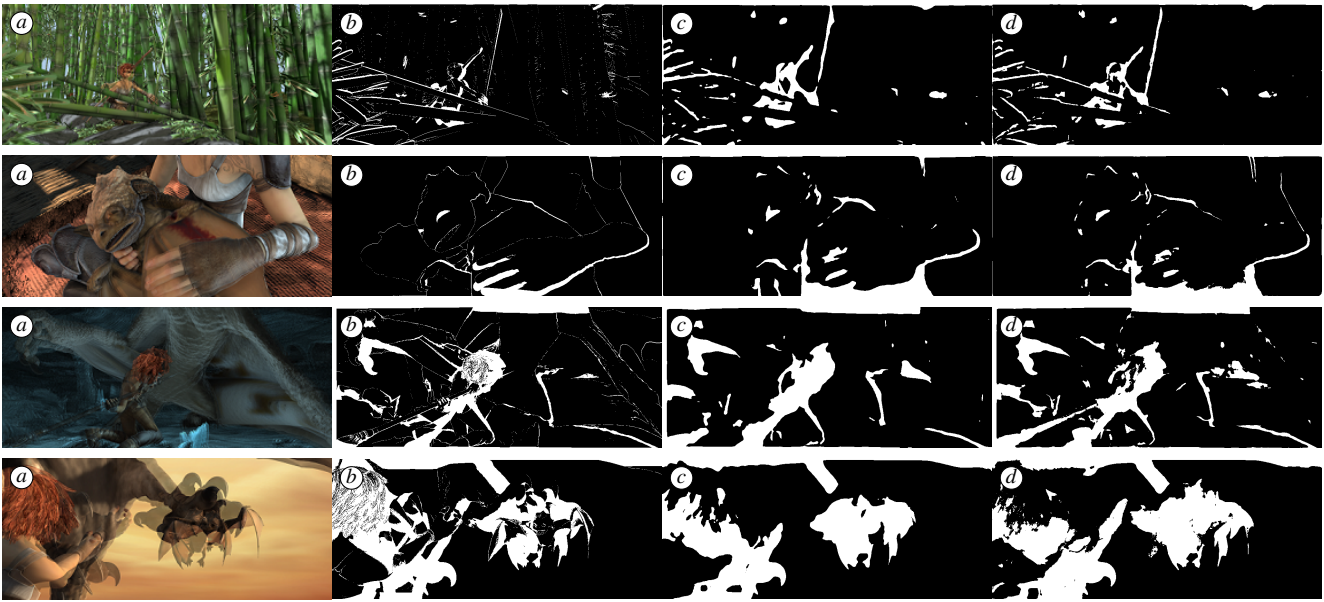
}


\section{Additional Qualitative Examples}
\label{sec:qualitative}

\paragraph{Occlusion upsampling layer.} \cref{fig:occ_ablation} provides qualitative examples of occlusion estimation and demonstrates the advantage of using the occlusion upsampling layer. 
The models used here are trained on the FlyingChairsOcc dataset only (no fine-tuning on the FlyingThings3D-subset dataset or Sintel) and tested on Sintel Train Clean.
The occlusion upsampling layer enhances the occlusion estimates to be much sharper along motion boundaries and refines coarse estimates. 
Also, the upsampling layer further detects thinly-shaped occlusions that were not detected at the quarter resolution. 
Unlike optical flow, where a quarter resolution estimate is largely sufficient, we can see from these qualitative examples that estimating occlusions up to the original resolution is very critical for yielding high accuracy.

\myparagraph{Ablation study on PWC-Net.}
In addition to \cref{fig:pwc_ablation} in the main paper, we here give more qualitative examples for the ablation study.
In \cref{fig:pwc_ablation_more}, all models are also trained on the FlyingChairsOcc dataset and tested on Sintel Train Clean.
Our proposed schemes significantly improve the accuracy over the baseline model (\ie PWC-Net \cite{Sun:2017:PWC}), yielding better generalization across datasets.

{
\begin{figure*}[ht]
\centering
\footnotesize
\color{black}
\setlength\tabcolsep{0.3pt}
\renewcommand{\arraystretch}{0.2}
\begin{tabular}{>{\centering\arraybackslash}m{.20\textwidth} >{\centering\arraybackslash}m{.20\textwidth} >{\centering\arraybackslash}m{.20\textwidth} >{\centering\arraybackslash}m{.20\textwidth} >{\centering\arraybackslash}m{.20\textwidth}}
	
	\begin{tikzpicture}
    \node[inner sep=0] (img) {\includegraphics[width=\linewidth]{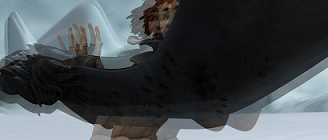}};    
    \node[anchor=north west] at (img.north west){\circled{a}};
	\end{tikzpicture}&
	\begin{tikzpicture}
    \node[inner sep=0] (img) {\includegraphics[width=\linewidth]{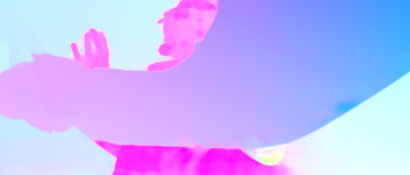}};    
    \node[anchor=north west] at (img.north west){\circled{b}};
	\end{tikzpicture}&
	\begin{tikzpicture}
    \node[inner sep=0] (img) {\includegraphics[width=\linewidth]{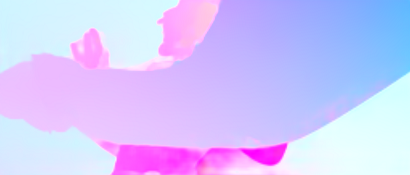}};    
    \node[anchor=north west] at (img.north west){\circled{c}};
	\end{tikzpicture}&
	\begin{tikzpicture}
    \node[inner sep=0] (img) {\includegraphics[width=\linewidth]{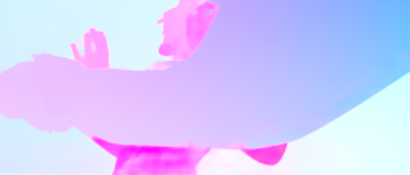}};    
    \node[anchor=north west] at (img.north west){\circled{d}};
	\end{tikzpicture}&
	\begin{tikzpicture}
    \node[inner sep=0] (img) {\includegraphics[width=\linewidth]{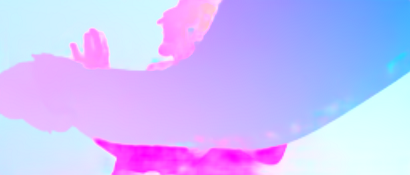}};    
    \node[anchor=north west] at (img.north west){\circled{e}};
	\end{tikzpicture}\\	
	\begin{tikzpicture}
    \node[inner sep=0] (img) {\includegraphics[width=\linewidth]{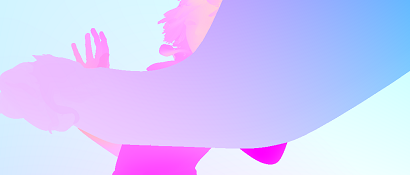}};    
    \node[anchor=north west] at (img.north west){\circled{f}};
	\end{tikzpicture}&
	\begin{tikzpicture}
    \node[inner sep=0] (img) {\includegraphics[width=\linewidth]{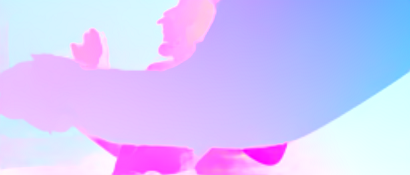}};    
    \node[anchor=north west] at (img.north west){\circled{g}};
	\end{tikzpicture}&
	\begin{tikzpicture}
    \node[inner sep=0] (img) {\includegraphics[width=\linewidth]{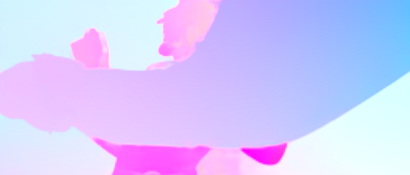}};    
    \node[anchor=north west] at (img.north west){\circled{h}};
	\end{tikzpicture}&
	\begin{tikzpicture}
    \node[inner sep=0] (img) {\includegraphics[width=\linewidth]{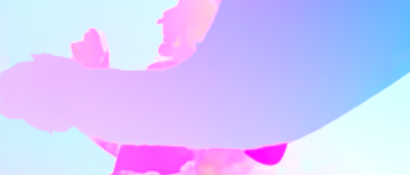}};    
    \node[anchor=north west] at (img.north west){\circled{i}};
	\end{tikzpicture}&
	\begin{tikzpicture}
    \node[inner sep=0] (img) {\includegraphics[width=\linewidth]{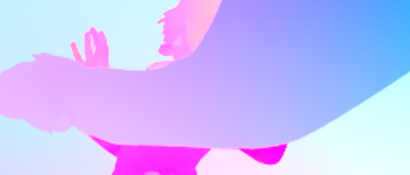}};    
    \node[anchor=north west] at (img.north west){\circled{j}};
	\end{tikzpicture}\\  \\

	\begin{tikzpicture}
    \node[inner sep=0] (img) {\includegraphics[width=\linewidth]{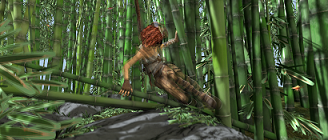}};    
    \node[anchor=north west] at (img.north west){\circled{a}};
	\end{tikzpicture}&
	\begin{tikzpicture}
    \node[inner sep=0] (img) {\includegraphics[width=\linewidth]{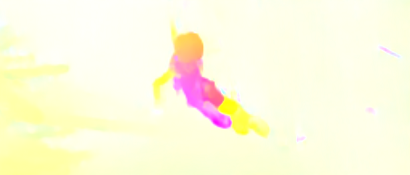}};    
    \node[anchor=north west] at (img.north west){\circled{b}};
	\end{tikzpicture}&
	\begin{tikzpicture}
    \node[inner sep=0] (img) {\includegraphics[width=\linewidth]{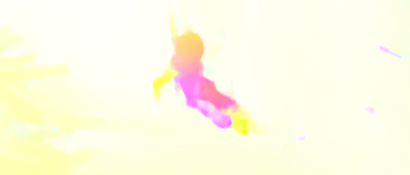}};    
    \node[anchor=north west] at (img.north west){\circled{c}};
	\end{tikzpicture}&
	\begin{tikzpicture}
    \node[inner sep=0] (img) {\includegraphics[width=\linewidth]{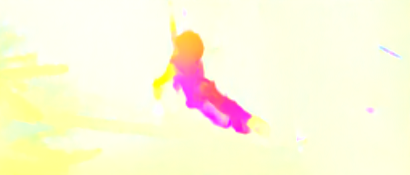}};    
    \node[anchor=north west] at (img.north west){\circled{d}};
	\end{tikzpicture}&
	\begin{tikzpicture}
    \node[inner sep=0] (img) {\includegraphics[width=\linewidth]{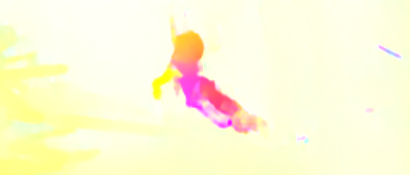}};    
    \node[anchor=north west] at (img.north west){\circled{e}};
	\end{tikzpicture}\\	
	\begin{tikzpicture}
    \node[inner sep=0] (img) {\includegraphics[width=\linewidth]{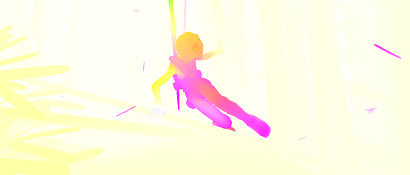}};    
    \node[anchor=north west] at (img.north west){\circled{f}};
	\end{tikzpicture}&
	\begin{tikzpicture}
    \node[inner sep=0] (img) {\includegraphics[width=\linewidth]{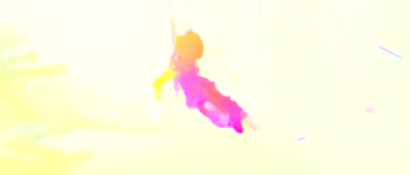}};    
    \node[anchor=north west] at (img.north west){\circled{g}};
	\end{tikzpicture}&
	\begin{tikzpicture}
    \node[inner sep=0] (img) {\includegraphics[width=\linewidth]{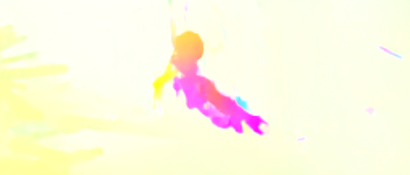}};    
    \node[anchor=north west] at (img.north west){\circled{h}};
	\end{tikzpicture}&
	\begin{tikzpicture}
    \node[inner sep=0] (img) {\includegraphics[width=\linewidth]{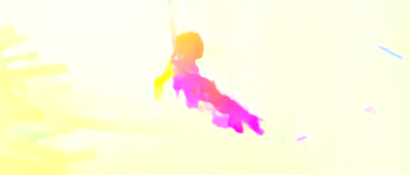}};    
    \node[anchor=north west] at (img.north west){\circled{i}};
	\end{tikzpicture}&
	\begin{tikzpicture}
    \node[inner sep=0] (img) {\includegraphics[width=\linewidth]{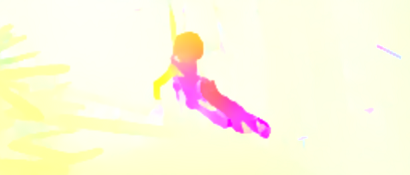}};    
    \node[anchor=north west] at (img.north west){\circled{j}};
	\end{tikzpicture}\\  \\
	
	\begin{tikzpicture}
    \node[inner sep=0] (img) {\includegraphics[width=\linewidth]{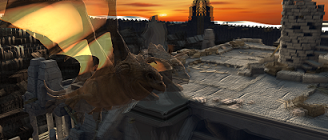}};    
    \node[anchor=north west] at (img.north west){\circled{a}};
	\end{tikzpicture}&
	\begin{tikzpicture}
    \node[inner sep=0] (img) {\includegraphics[width=\linewidth]{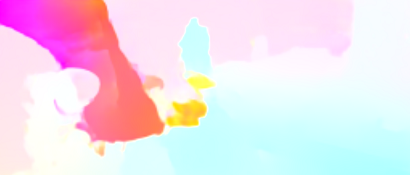}};    
    \node[anchor=north west] at (img.north west){\circled{b}};
	\end{tikzpicture}&
	\begin{tikzpicture}
    \node[inner sep=0] (img) {\includegraphics[width=\linewidth]{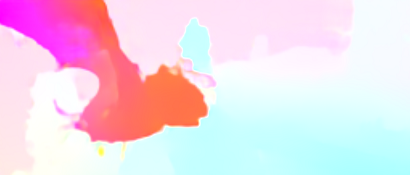}};    
    \node[anchor=north west] at (img.north west){\circled{c}};
	\end{tikzpicture}&
	\begin{tikzpicture}
    \node[inner sep=0] (img) {\includegraphics[width=\linewidth]{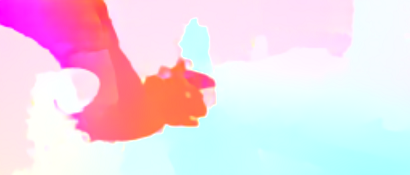}};    
    \node[anchor=north west] at (img.north west){\circled{d}};
	\end{tikzpicture}&
	\begin{tikzpicture}
    \node[inner sep=0] (img) {\includegraphics[width=\linewidth]{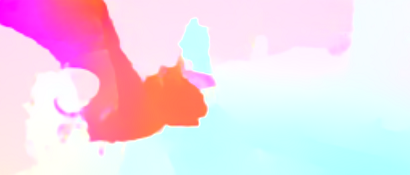}};    
    \node[anchor=north west] at (img.north west){\circled{e}};
	\end{tikzpicture}\\	
	\begin{tikzpicture}
    \node[inner sep=0] (img) {\includegraphics[width=\linewidth]{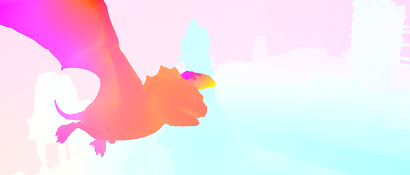}};    
    \node[anchor=north west] at (img.north west){\circled{f}};
	\end{tikzpicture}&
	\begin{tikzpicture}
    \node[inner sep=0] (img) {\includegraphics[width=\linewidth]{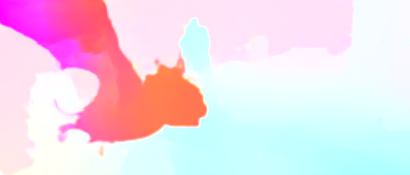}};    
    \node[anchor=north west] at (img.north west){\circled{g}};
	\end{tikzpicture}&
	\begin{tikzpicture}
    \node[inner sep=0] (img) {\includegraphics[width=\linewidth]{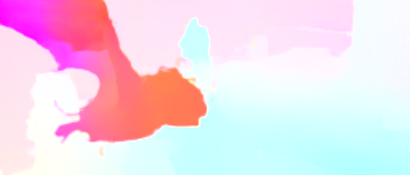}};    
    \node[anchor=north west] at (img.north west){\circled{h}};
	\end{tikzpicture}&
	\begin{tikzpicture}
    \node[inner sep=0] (img) {\includegraphics[width=\linewidth]{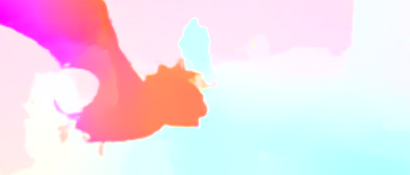}};    
    \node[anchor=north west] at (img.north west){\circled{i}};
	\end{tikzpicture}&
	\begin{tikzpicture}
    \node[inner sep=0] (img) {\includegraphics[width=\linewidth]{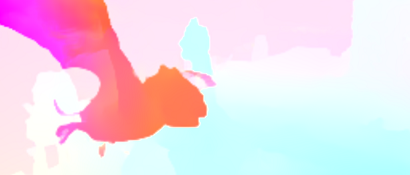}};    
    \node[anchor=north west] at (img.north west){\circled{j}};
	\end{tikzpicture}\\ \\
	
	\begin{tikzpicture}
    \node[inner sep=0] (img) {\includegraphics[width=\linewidth]{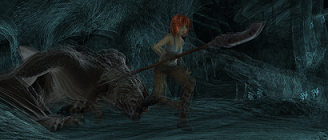}};    
    \node[anchor=north west] at (img.north west){\circled{a}};
	\end{tikzpicture}&
	\begin{tikzpicture}
    \node[inner sep=0] (img) {\includegraphics[width=\linewidth]{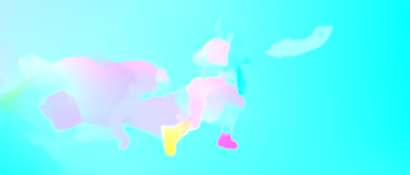}};    
    \node[anchor=north west] at (img.north west){\circled{b}};
	\end{tikzpicture}&
	\begin{tikzpicture}
    \node[inner sep=0] (img) {\includegraphics[width=\linewidth]{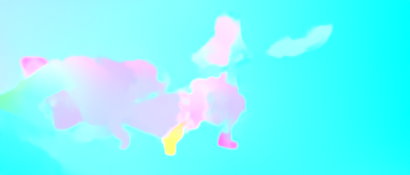}};    
    \node[anchor=north west] at (img.north west){\circled{c}};
	\end{tikzpicture}&
	\begin{tikzpicture}
    \node[inner sep=0] (img) {\includegraphics[width=\linewidth]{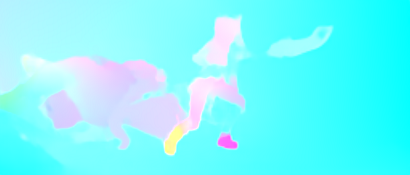}};    
    \node[anchor=north west] at (img.north west){\circled{d}};
	\end{tikzpicture}&
	\begin{tikzpicture}
    \node[inner sep=0] (img) {\includegraphics[width=\linewidth]{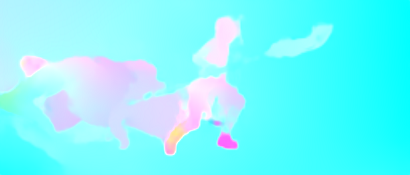}};    
    \node[anchor=north west] at (img.north west){\circled{e}};
	\end{tikzpicture}\\	
	\begin{tikzpicture}
    \node[inner sep=0] (img) {\includegraphics[width=\linewidth]{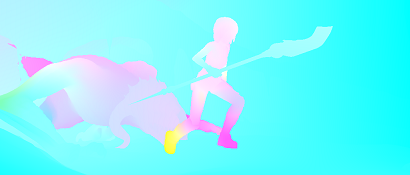}};    
    \node[anchor=north west] at (img.north west){\circled{f}};
	\end{tikzpicture}&
	\begin{tikzpicture}
    \node[inner sep=0] (img) {\includegraphics[width=\linewidth]{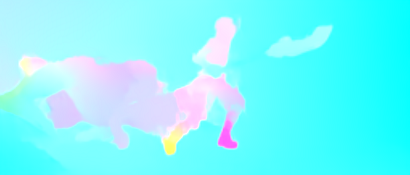}};    
    \node[anchor=north west] at (img.north west){\circled{g}};
	\end{tikzpicture}&
	\begin{tikzpicture}
    \node[inner sep=0] (img) {\includegraphics[width=\linewidth]{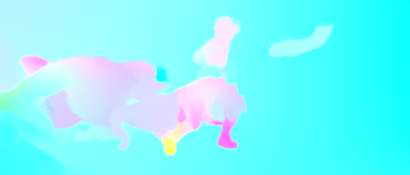}};    
    \node[anchor=north west] at (img.north west){\circled{h}};
	\end{tikzpicture}&
	\begin{tikzpicture}
    \node[inner sep=0] (img) {\includegraphics[width=\linewidth]{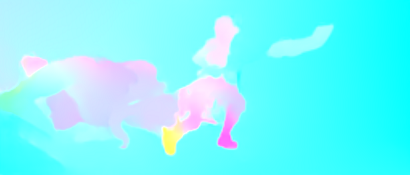}};    
    \node[anchor=north west] at (img.north west){\circled{i}};
	\end{tikzpicture}&
	\begin{tikzpicture}
    \node[inner sep=0] (img) {\includegraphics[width=\linewidth]{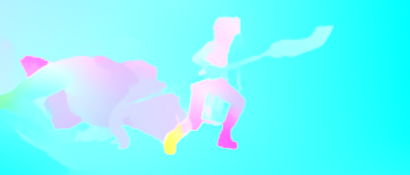}};    
    \node[anchor=north west] at (img.north west){\circled{j}};
	\end{tikzpicture}\\ \\
\end{tabular}
\caption {\textbf{More qualitative examples from the ablation study on PWC-Net}: \emph{(a)} overlapped input images, \emph{(b)} the original PWC-Net \cite{Sun:2017:PWC}, \emph{(c)} PWC-Net with Bi, \emph{(d)} PWC-Net with Occ, \emph{(e)} PWC-Net with Bi-Occ, \emph{(f)} optical flow ground truth, \emph{(g)} PWC-Net with IRR, \emph{(h)} PWC-Net with Occ-IRR, \emph{(i)} PWC-Net with Bi-Occ-IRR, and \emph{(j)} our full model (i.e. IRR-PWC). Our full model significantly improves flow estimation over the original PWC-Net with fewer missing details and clearer motion boundaries. Note that there are gradual improvements when combining several of the proposed components.}
\label{fig:pwc_ablation_more}
\end{figure*}
}


\section{Qualitative Comparison}
\label{sec:comparison}

\subsection{Occlusion estimation}

\Cref{fig:occ_compare_supp} demonstrates a qualitative comparison with the state of the art on occlusion estimation. 
Qualitatively, MirrorFlow \cite{Hur:2017:MFE} misses many occlusions in general, and FlowNet-CSSR-ft-sd \cite{Ilg:2018:OMD} is able to detect fine details of occlusion.
In contrast, our method tries not to miss occlusions, which results in a better recall rate but somewhat lower precision than those of FlowNet-CSSR-ft-sd \cite{Ilg:2018:OMD}.
Overall, our method demonstrates better F1-score than FlowNet-CSSR-ft-sd \cite{Ilg:2018:OMD}, achieving state-of-the-art results on the evaluation dataset (\ie Sintel Train Clean and Final).
Note that FlowNet-CSSR-ft-sd \cite{Ilg:2018:OMD} is additionally trained on the ChairsSDHom dataset \cite{Ilg:2017:FN2} for handling small-displacement motion, which is related to thinly-shaped occlusions.
Our approach is not trained further.

{
\begin{figure*}[!t]
\centering
\footnotesize
\setlength\tabcolsep{0.3pt}
\renewcommand{\arraystretch}{0.2}
\begin{tabular}{>{\centering\arraybackslash}m{.20\textwidth} >{\centering\arraybackslash}m{.20\textwidth} >{\centering\arraybackslash}m{.20\textwidth} >{\centering\arraybackslash}m{.20\textwidth} >{\centering\arraybackslash}m{.20\textwidth}}
	
	\begin{tikzpicture}
    \node[inner sep=0] (img) {\includegraphics[width=\linewidth]{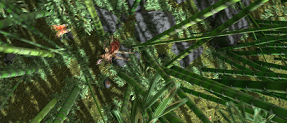}};    
    \node[anchor=north west] at (img.north west){\circled{a}};
	\end{tikzpicture}&
	\begin{tikzpicture}
    \node[inner sep=0] (img) {\includegraphics[width=\linewidth]{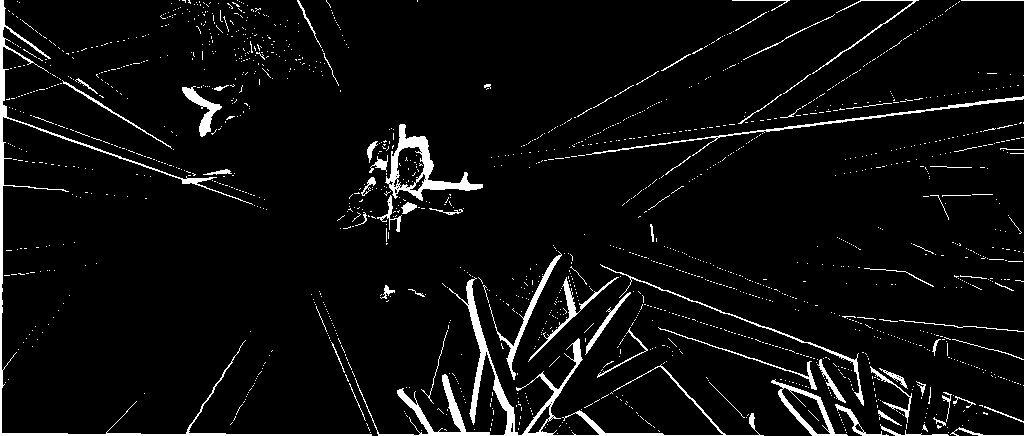}};    
    \node[anchor=north west] at (img.north west){\circled{b}};
	\end{tikzpicture}&
	\begin{tikzpicture}
    \node[inner sep=0] (img) {\includegraphics[width=\linewidth]{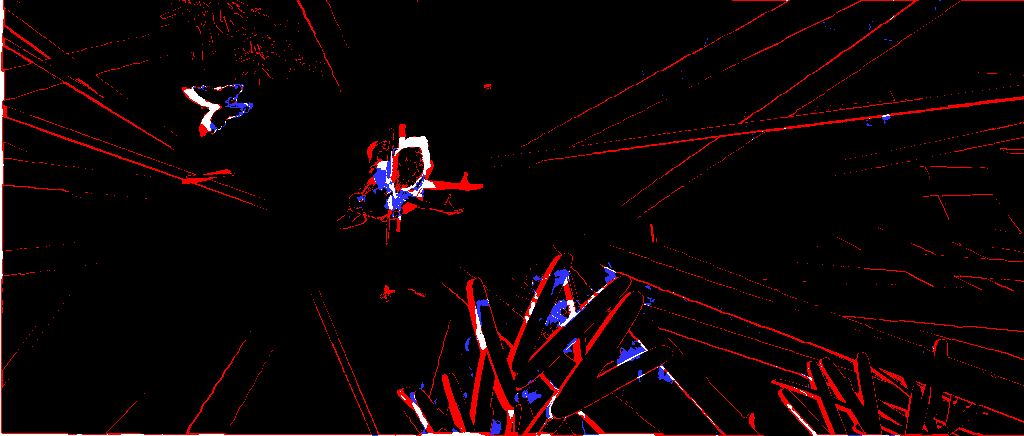}};    
    \node[anchor=north west] at (img.north west){\circled{c}};
    \node[anchor=north east] at (img.north east){\tiny \color{white} \textbf{F-score: 0.281}};
	\end{tikzpicture}&
	\begin{tikzpicture}
    \node[inner sep=0] (img) {\includegraphics[width=\linewidth]{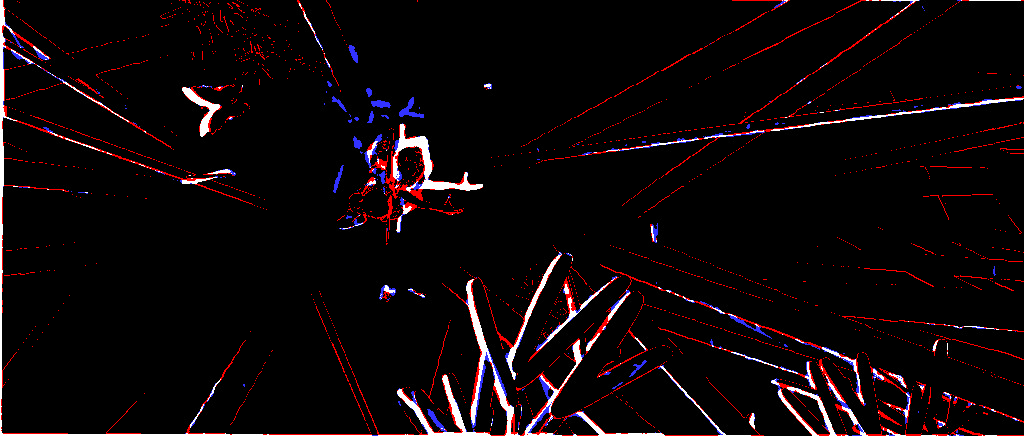}};    
    \node[anchor=north west] at (img.north west){\circled{d}};
    \node[anchor=north east] at (img.north east){\tiny \color{white}\textbf{F-score: 0.579}};
	\end{tikzpicture}&
	\begin{tikzpicture}
	\node[inner sep=0] (img) {\includegraphics[width=\linewidth]{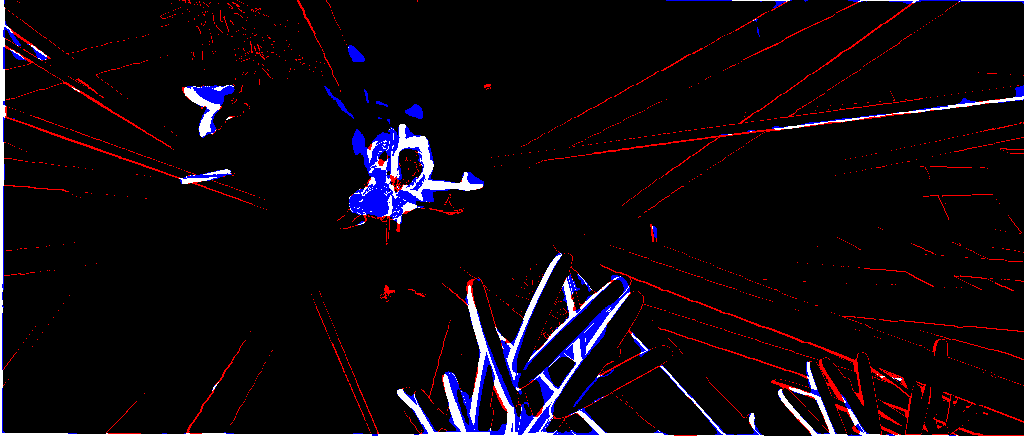}};    
    \node[anchor=north west] at (img.north west){\circled{e}};
    \node[anchor=north east] at (img.north east){\tiny \color{white} \textbf{F-score: 0.541}};
	\end{tikzpicture} \\ \\

	\begin{tikzpicture}
    \node[inner sep=0] (img) {\includegraphics[width=\linewidth]{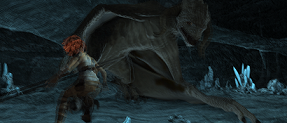}};    
    \node[anchor=north west] at (img.north west){\circled{a}};
	\end{tikzpicture}&
	\begin{tikzpicture}
    \node[inner sep=0] (img) {\includegraphics[width=\linewidth]{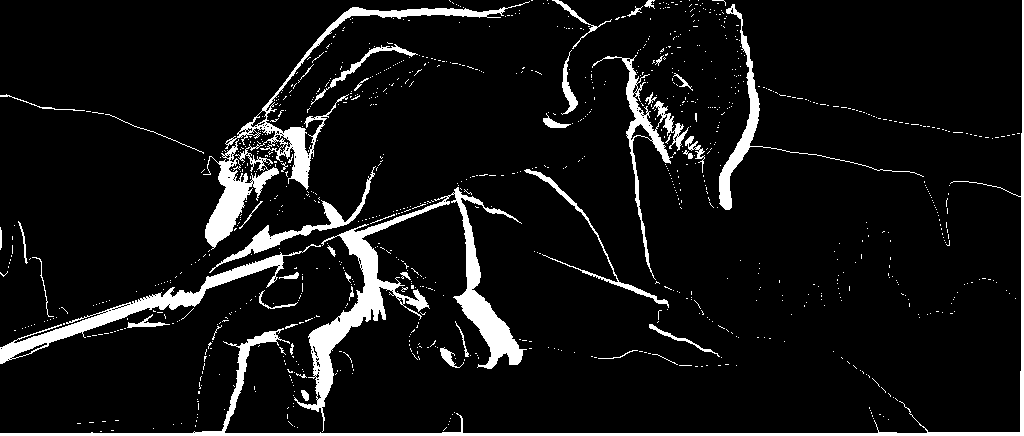}};    
    \node[anchor=north west] at (img.north west){\circled{b}};
	\end{tikzpicture}&
	\begin{tikzpicture}
    \node[inner sep=0] (img) {\includegraphics[width=\linewidth]{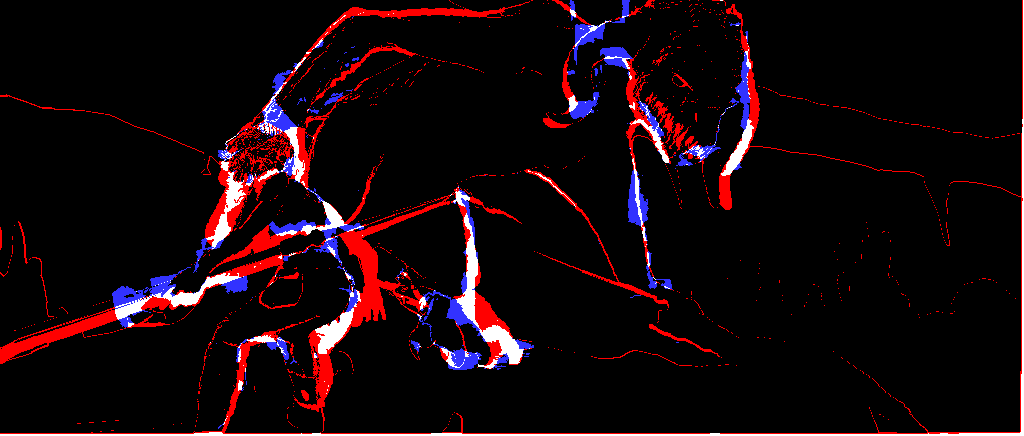}};    
    \node[anchor=north west] at (img.north west){\circled{c}};
    \node[anchor=north east] at (img.north east){\tiny \color{white} \textbf{F-score: 0.408}};
	\end{tikzpicture}&
	\begin{tikzpicture}
    \node[inner sep=0] (img) {\includegraphics[width=\linewidth]{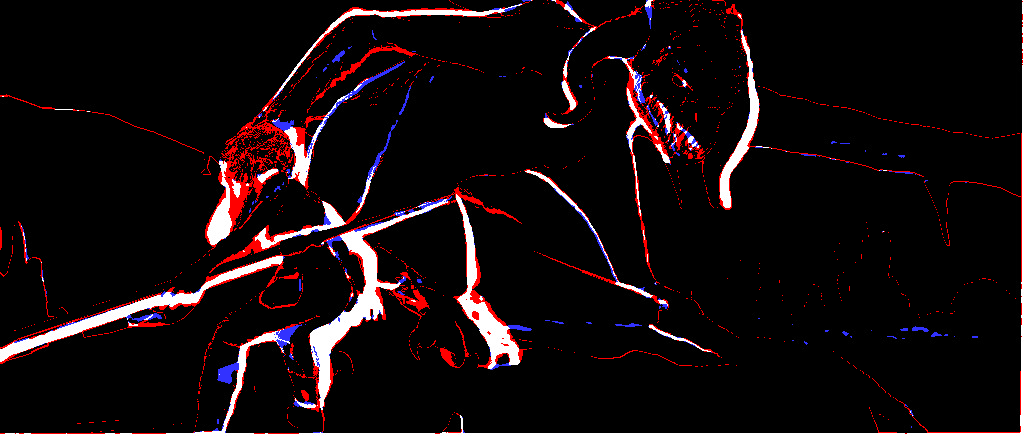}};    
    \node[anchor=north west] at (img.north west){\circled{d}};
    \node[anchor=north east] at (img.north east){\tiny \color{white} \textbf{F-score: 0.688}};
	\end{tikzpicture}&
	\begin{tikzpicture}
	\node[inner sep=0] (img) {\includegraphics[width=\linewidth]{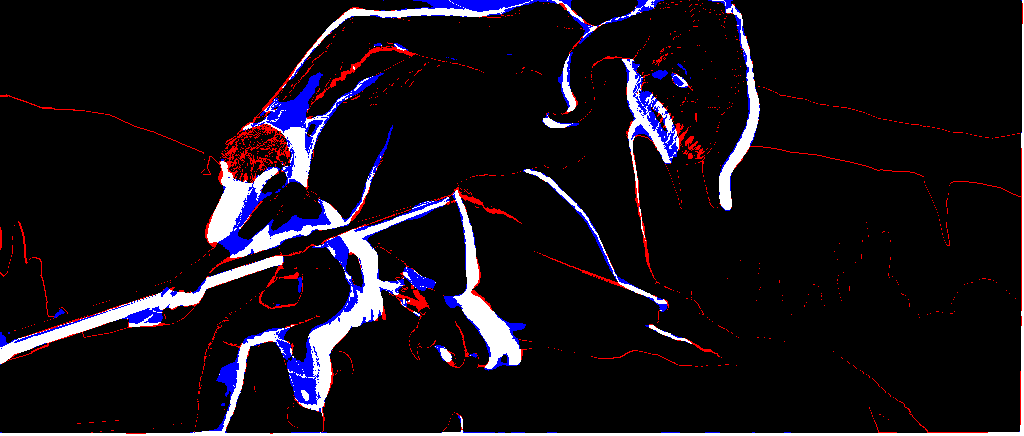}};    
    \node[anchor=north west] at (img.north west){\circled{e}};
    \node[anchor=north east] at (img.north east){\tiny \color{white} \textbf{F-score: 0.708}};
	\end{tikzpicture} \\ \\
	
	\begin{tikzpicture}
    \node[inner sep=0] (img) {\includegraphics[width=\linewidth]{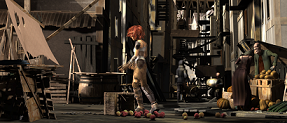}};    
    \node[anchor=north west] at (img.north west){\circled{a}};
	\end{tikzpicture}&
	\begin{tikzpicture}
    \node[inner sep=0] (img) {\includegraphics[width=\linewidth]{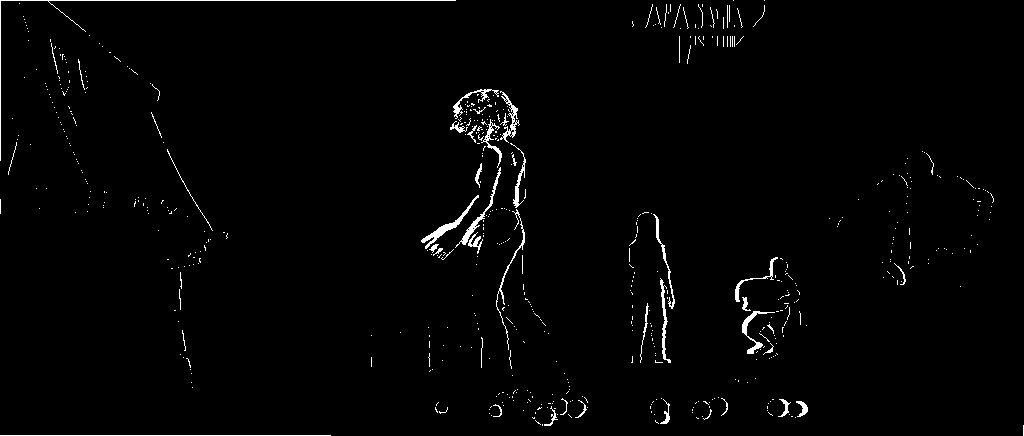}};    
    \node[anchor=north west] at (img.north west){\circled{b}};
	\end{tikzpicture}&
	\begin{tikzpicture}
    \node[inner sep=0] (img) {\includegraphics[width=\linewidth]{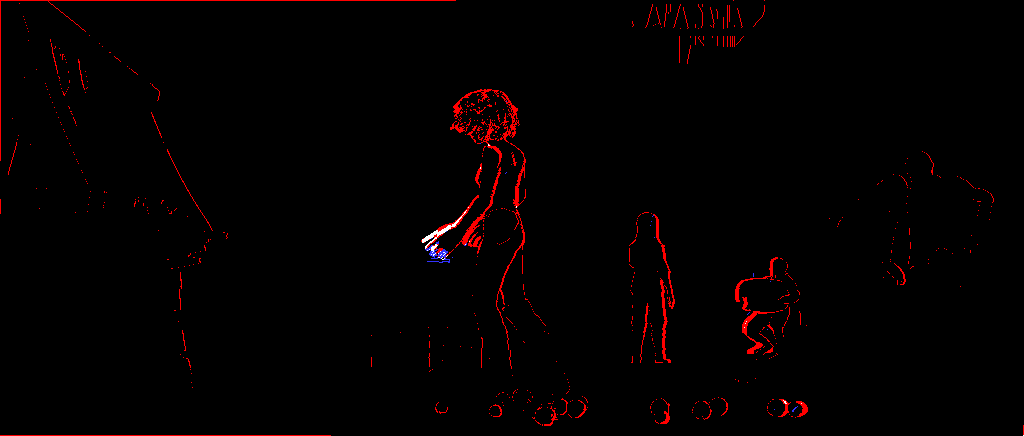}};    
    \node[anchor=north west] at (img.north west){\circled{c}};
    \node[anchor=north east] at (img.north east){\tiny \color{white} \textbf{F-score: 0.072}};
	\end{tikzpicture}&
	\begin{tikzpicture}
    \node[inner sep=0] (img) {\includegraphics[width=\linewidth]{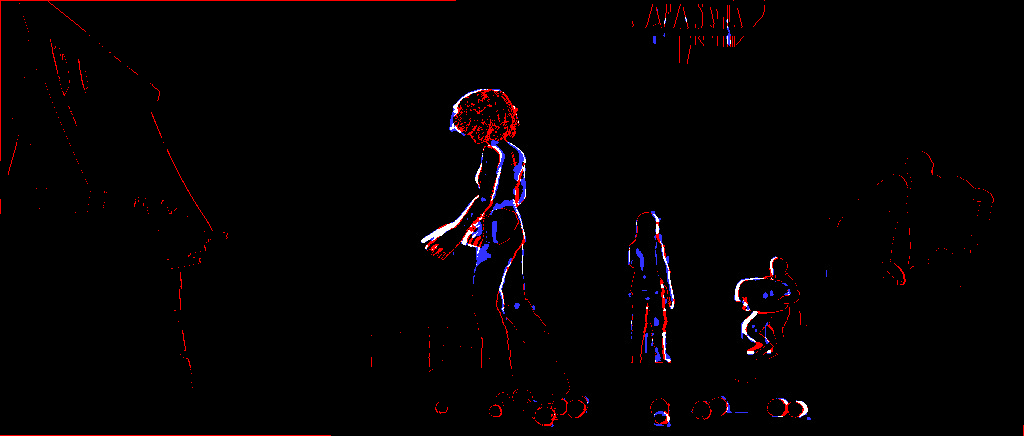}};    
    \node[anchor=north west] at (img.north west){\circled{d}};
    \node[anchor=north east] at (img.north east){\tiny \color{white} \textbf{F-score: 0.307}};
	\end{tikzpicture}&
	\begin{tikzpicture}
	\node[inner sep=0] (img) {\includegraphics[width=\linewidth]{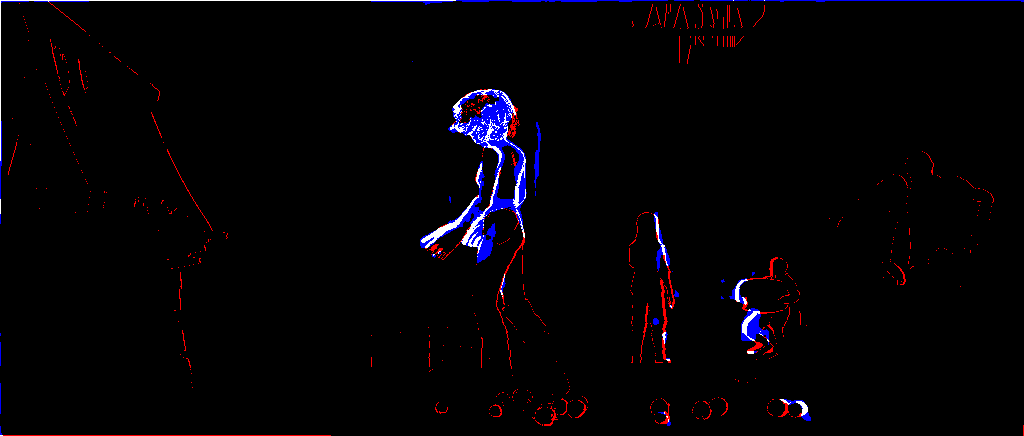}};    
    \node[anchor=north west] at (img.north west){\circled{e}};
    \node[anchor=north east] at (img.north east){\tiny \color{white} \textbf{F-score: 0.392}};
	\end{tikzpicture} \\ \\

\end{tabular}
\caption {\textbf{Qualitative comparison of occlusion estimation with the state of the art}: \emph{(a)} overlapped input images, \emph{(b)} occlusion ground truth, \emph{(c)} MirrorFlow \cite{Hur:2017:MFE}, \emph{(d)} FlowNet-CSSR-ft-sd \cite{Ilg:2018:OMD}, and \emph{(e)} ours. In the result image of each method, blue pixels denote \textbf{\color{blue}false positives}, red pixels denote \textbf{\color{red}false negatives}, and white ones denote true positives (\ie correctly estimated occlusion). We include the F-score of each method in the top-right corner. Our model yields a better F-score on the second and the third scene than FlowNet-CSSR-ft-sd \cite{Ilg:2018:OMD}.}
\label{fig:occ_compare_supp}
\end{figure*}
}

\subsection{Bi-directional flows and occlusion maps}

MirrorFlow \cite{Hur:2017:MFE} is one of the most recent related works that estimates bi-directional flow and occlusion maps.
\cref{fig:bidirection_comparison} provides a qualitative comparison with MirrorFlow \cite{Hur:2017:MFE} on the Sintel and KITTI 2015 datasets.
In this comparison, we use our model fine-tuned on the training set of each dataset and display qualitative examples from the validation split.
Comparing to MirrorFlow \cite{Hur:2017:MFE}, our model demonstrates far fewer artifacts and fewer missing details for both flow and occlusion estimation. 
Although there is no ground truth for backward flow nor an occlusion map for the second image available for supervision, our bi-directional model is able to estimate the backward flow and the second occlusion map well while only using the ground truth of forward flow and the occlusion map for the first image (latter is only available on Sintel) during fine-tuning.

{
\begin{figure*}[!b]
\centering
\scriptsize
\setlength\tabcolsep{0.3pt}
\renewcommand{\arraystretch}{0.2}
\begin{tabular}{>{\centering\arraybackslash}m{.20\textwidth} >{\centering\arraybackslash}m{.20\textwidth} >{\centering\arraybackslash}m{.20\textwidth} >{\centering\arraybackslash}m{.20\textwidth} >{\centering\arraybackslash}m{.20\textwidth}}

	\begin{tikzpicture}
    \node[inner sep=0] (img) {\includegraphics[width=\linewidth]{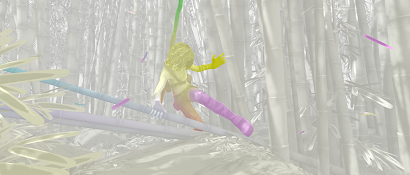}};    
    \node[anchor=north west] at (img.north west){\circled{a}};
	\end{tikzpicture}&
	\begin{tikzpicture}
    \node[inner sep=0] (img) {\includegraphics[width=\linewidth]{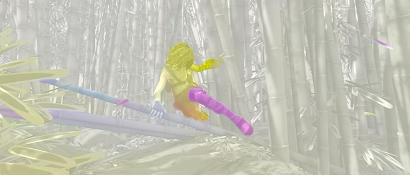}};    
    \node[anchor=north west] at (img.north west){\circled{b}};
	\end{tikzpicture}&
	\begin{tikzpicture}
    \node[inner sep=0] (img) {\includegraphics[width=\linewidth]{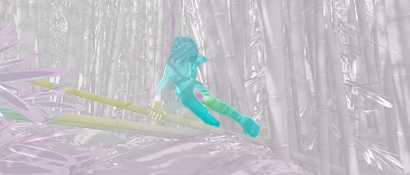}};    
    \node[anchor=north west] at (img.north west){\circled{c}};
	\end{tikzpicture}&
	\begin{tikzpicture}
    \node[inner sep=0] (img) {\includegraphics[width=\linewidth]{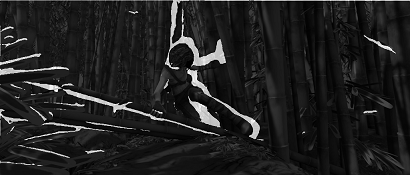}};    
    \node[anchor=north west] at (img.north west){\circled{d}};
	\end{tikzpicture}&
	\begin{tikzpicture}
	\node[inner sep=0] (img) {\includegraphics[width=\linewidth]{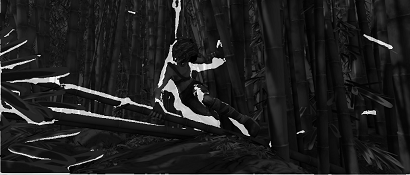}};    
    \node[anchor=north west] at (img.north west){\circled{e}};
	\end{tikzpicture} \\	
	
	\begin{tikzpicture}
    \node[inner sep=0] (img) {\includegraphics[width=\linewidth]{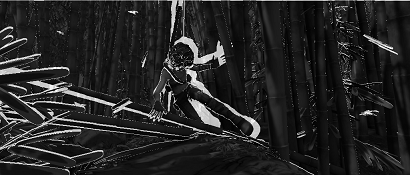}};    
    \node[anchor=north west] at (img.north west){\circled{f}};
	\end{tikzpicture}&
	\begin{tikzpicture}
    \node[inner sep=0] (img) {\includegraphics[width=\linewidth]{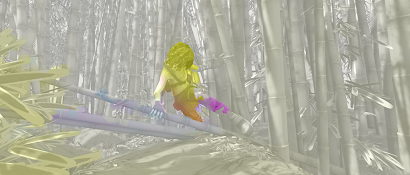}};    
    \node[anchor=north west] at (img.north west){\circled{g}};
	\end{tikzpicture}&
	\begin{tikzpicture}
    \node[inner sep=0] (img) {\includegraphics[width=\linewidth]{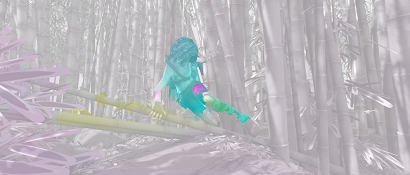}};    
    \node[anchor=north west] at (img.north west){\circled{h}};
	\end{tikzpicture}&
	\begin{tikzpicture}
    \node[inner sep=0] (img) {\includegraphics[width=\linewidth]{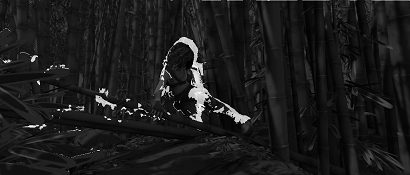}};    
    \node[anchor=north west] at (img.north west){\circled{i}};
	\end{tikzpicture}&
	\begin{tikzpicture}
	\node[inner sep=0] (img) {\includegraphics[width=\linewidth]{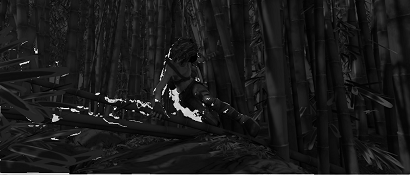}};    
    \node[anchor=north west] at (img.north west){\circled{j}};
	\end{tikzpicture} \\ \\

	\begin{tikzpicture}
    \node[inner sep=0] (img) {\includegraphics[width=\linewidth]{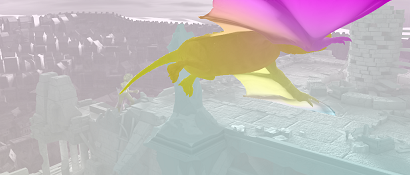}};    
    \node[anchor=north west] at (img.north west){\circled{a}};
	\end{tikzpicture}&
	\begin{tikzpicture}
    \node[inner sep=0] (img) {\includegraphics[width=\linewidth]{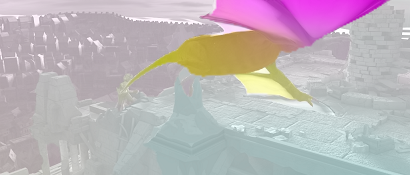}};    
    \node[anchor=north west] at (img.north west){\circled{b}};
	\end{tikzpicture}&
	\begin{tikzpicture}
    \node[inner sep=0] (img) {\includegraphics[width=\linewidth]{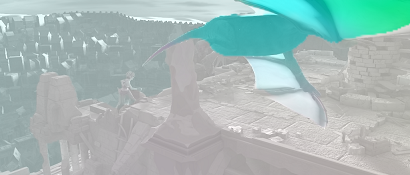}};    
    \node[anchor=north west] at (img.north west){\circled{c}};
	\end{tikzpicture}&
	\begin{tikzpicture}
    \node[inner sep=0] (img) {\includegraphics[width=\linewidth]{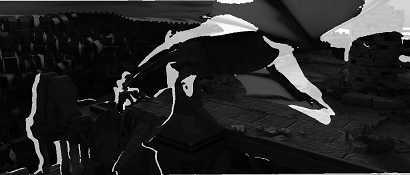}};    
    \node[anchor=north west] at (img.north west){\circled{d}};
	\end{tikzpicture}&
	\begin{tikzpicture}
	\node[inner sep=0] (img) {\includegraphics[width=\linewidth]{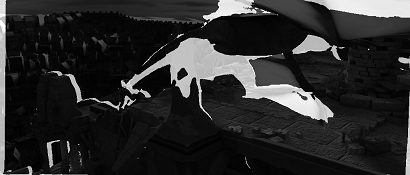}};    
    \node[anchor=north west] at (img.north west){\circled{e}};
	\end{tikzpicture} \\	

	\begin{tikzpicture}
    \node[inner sep=0] (img) {\includegraphics[width=\linewidth]{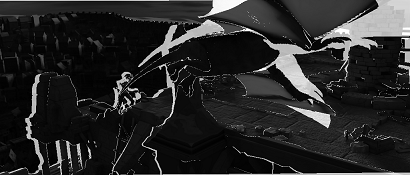}};    
    \node[anchor=north west] at (img.north west){\circled{f}};
	\end{tikzpicture}&
	\begin{tikzpicture}
    \node[inner sep=0] (img) {\includegraphics[width=\linewidth]{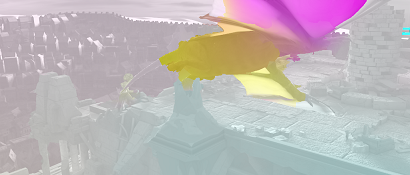}};    
    \node[anchor=north west] at (img.north west){\circled{g}};
	\end{tikzpicture}&
	\begin{tikzpicture}
    \node[inner sep=0] (img) {\includegraphics[width=\linewidth]{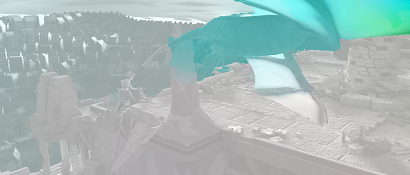}};    
    \node[anchor=north west] at (img.north west){\circled{h}};
	\end{tikzpicture}&
	\begin{tikzpicture}
    \node[inner sep=0] (img) {\includegraphics[width=\linewidth]{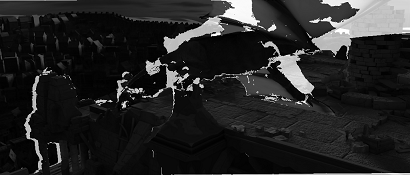}};    
    \node[anchor=north west] at (img.north west){\circled{i}};
	\end{tikzpicture}&
	\begin{tikzpicture}
	\node[inner sep=0] (img) {\includegraphics[width=\linewidth]{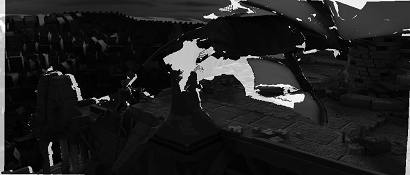}};    
    \node[anchor=north west] at (img.north west){\circled{j}};
	\end{tikzpicture} \\ \\
		
	\begin{tikzpicture}
    \node[inner sep=0] (img) {\includegraphics[width=\linewidth]{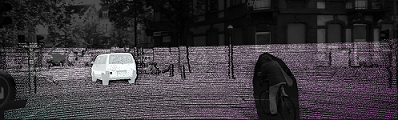}};    
    \node[anchor=north west] at (img.north west){\circled{a}};
	\end{tikzpicture}&
	\begin{tikzpicture}
    \node[inner sep=0] (img) {\includegraphics[width=\linewidth]{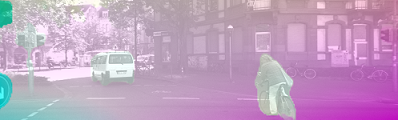}};    
    \node[anchor=north west] at (img.north west){\circled{b}};
	\end{tikzpicture}&
	\begin{tikzpicture}
    \node[inner sep=0] (img) {\includegraphics[width=\linewidth]{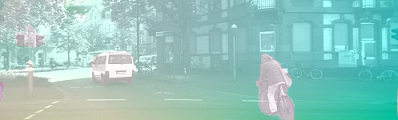}};    
    \node[anchor=north west] at (img.north west){\circled{c}};
	\end{tikzpicture}&
	\begin{tikzpicture}
    \node[inner sep=0] (img) {\includegraphics[width=\linewidth]{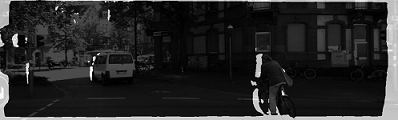}};    
    \node[anchor=north west] at (img.north west){\circled{d}};
	\end{tikzpicture}&
	\begin{tikzpicture}
	\node[inner sep=0] (img) {\includegraphics[width=\linewidth]{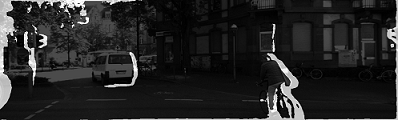}};    
    \node[anchor=north west] at (img.north west){\circled{e}};
	\end{tikzpicture} \\	
	&
	\begin{tikzpicture}
    \node[inner sep=0] (img) {\includegraphics[width=\linewidth]{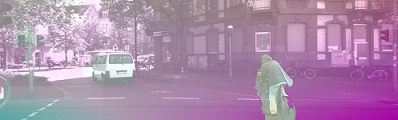}};    
    \node[anchor=north west] at (img.north west){\circled{g}};
	\end{tikzpicture}&
	\begin{tikzpicture}
    \node[inner sep=0] (img) {\includegraphics[width=\linewidth]{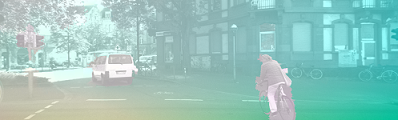}};    
    \node[anchor=north west] at (img.north west){\circled{h}};
	\end{tikzpicture}&
	\begin{tikzpicture}
    \node[inner sep=0] (img) {\includegraphics[width=\linewidth]{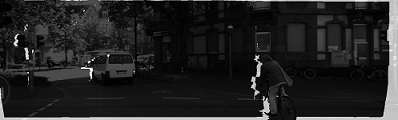}};    
    \node[anchor=north west] at (img.north west){\circled{i}};
	\end{tikzpicture}&
	\begin{tikzpicture}
	\node[inner sep=0] (img) {\includegraphics[width=\linewidth]{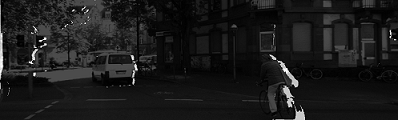}};    
    \node[anchor=north west] at (img.north west){\circled{j}};
	\end{tikzpicture} \\ \\
	
	\begin{tikzpicture}
    \node[inner sep=0] (img) {\includegraphics[width=\linewidth]{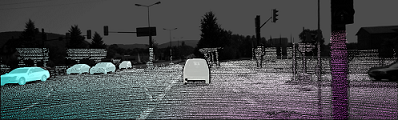}};    
    \node[anchor=north west] at (img.north west){\circled{a}};
	\end{tikzpicture}&
	\begin{tikzpicture}
    \node[inner sep=0] (img) {\includegraphics[width=\linewidth]{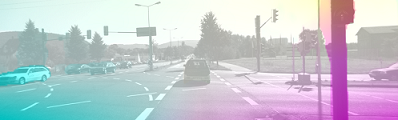}};    
    \node[anchor=north west] at (img.north west){\circled{b}};
	\end{tikzpicture}&
	\begin{tikzpicture}
    \node[inner sep=0] (img) {\includegraphics[width=\linewidth]{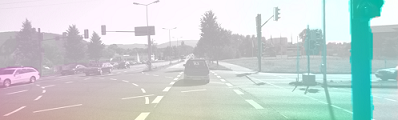}};    
    \node[anchor=north west] at (img.north west){\circled{c}};
	\end{tikzpicture}&
	\begin{tikzpicture}
    \node[inner sep=0] (img) {\includegraphics[width=\linewidth]{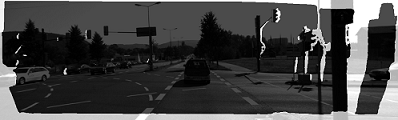}};    
    \node[anchor=north west] at (img.north west){\circled{d}};
	\end{tikzpicture}&
	\begin{tikzpicture}
	\node[inner sep=0] (img) {\includegraphics[width=\linewidth]{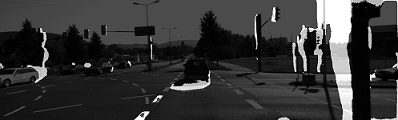}};    
    \node[anchor=north west] at (img.north west){\circled{e}};
	\end{tikzpicture} \\	
	&
	\begin{tikzpicture}
    \node[inner sep=0] (img) {\includegraphics[width=\linewidth]{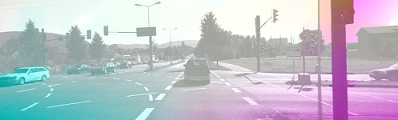}};    
    \node[anchor=north west] at (img.north west){\circled{g}};
	\end{tikzpicture}&
	\begin{tikzpicture}
    \node[inner sep=0] (img) {\includegraphics[width=\linewidth]{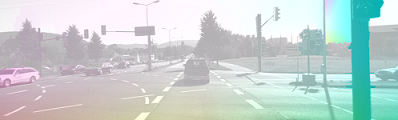}};    
    \node[anchor=north west] at (img.north west){\circled{h}};
	\end{tikzpicture}&
	\begin{tikzpicture}
    \node[inner sep=0] (img) {\includegraphics[width=\linewidth]{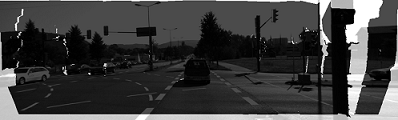}};    
    \node[anchor=north west] at (img.north west){\circled{i}};
	\end{tikzpicture}&
	\begin{tikzpicture}
	\node[inner sep=0] (img) {\includegraphics[width=\linewidth]{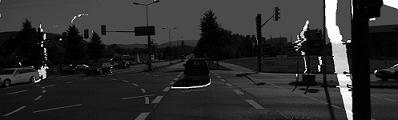}};    
    \node[anchor=north west] at (img.north west){\circled{j}};
	\end{tikzpicture} \\ \\

\end{tabular}
\caption {\textbf{Qualitative comparison of the bi-drectional optical flows and occlusion maps in both views with MirrorFlow \cite{Hur:2017:MFE}}: All results are overlayed on the corresponding image, either the first frame or the second frame. \emph{(a)} Ground truth optical flow, \emph{(b)} our forward flow, \emph{(c)} our backward flow, \emph{(d)} our occlusion map for the first frame, \emph{(e)} our occlusion map for the second frame, \emph{(f)} ground truth occlusion map, \emph{(g)} forward flow of MirrorFlow, \emph{(h)} backward flow of MirrorFlow, \emph{(i)} occlusion map of MirrorFlow for the first frame, \emph{(j)} occlusion map of MirrorFlow for the second frame. Note that KITTI has only sparse ground truth for optical flow and does not provide ground truth for occlusion.}
\label{fig:bidirection_comparison}
\end{figure*}
}

\end{document}